\documentclass[num-refs]{wiley-article}

\usepackage{siunitx}
\usepackage{hyperref}
\usepackage{graphicx}
\usepackage[table]{xcolor}
\usepackage{wrapfig}
\usepackage{longtable}
\usepackage{multirow}
\usepackage{colortbl}

\usepackage{soul}
\usepackage{color}

\newcommand{\hlc}[2][yellow]{ {\sethlcolor{#1} \hl{#2}} }

\newcolumntype{P}[1]{>{\centering\arraybackslash}p{#1}}
\newcolumntype{M}[1]{>{\centering\arraybackslash}m{#1}}

\definecolor{darkblue}{rgb}{0, 0, 0.5}
\hypersetup{colorlinks=true,citecolor=darkblue, linkcolor=darkblue, urlcolor=darkblue}

\papertype{Original Article}
\paperfield{Journal Section}

\title{The Unreasonable Effectiveness of Machine Learning in Moldavian versus Romanian Dialect Identification}

\author[1\authfn{1}]{Mihaela G\u{a}man}
\author[1,2\authfn{1}]{Radu Tudor Ionescu}

\contrib[\authfn{1}]{Equally contributing authors.}

\affil[1]{Department of Computer Science, University of Bucharest, 14 Academiei, Bucharest, 010014, Romania}
\affil[2]{Romanian Young Academy, University of Bucharest, 90 Panduri, Bucharest, 050663, Romania}

\corraddress{Radu Tudor Ionescu, Department of Computer Science, University of Bucharest, 14 Academiei, Bucharest, 010014, Romania}
\corremail{raducu.ionescu@gmail.com}

\fundinginfo{This work was supported by a grant of the Romanian Ministry of Education and Research, CNCS - UEFISCDI, project number PN-III-P1-1.1-TE-2019-0235, within PNCDI III. This article has also benefited from the support of the Romanian Young Academy, which is funded by Stiftung Mercator and the Alexander von Humboldt Foundation for the period 2020-2022.}

\runningauthor{G\u{a}man and Ionescu}

\begin{document}

\definecolor{red1}{rgb}{1.0,0.0,0.0}
\definecolor{red2}{rgb}{1.0,0.1,0.1}
\definecolor{red3}{rgb}{1.0,0.2,0.2}
\definecolor{red4}{rgb}{1.0,0.3,0.3}
\definecolor{red5}{rgb}{1.0,0.4,0.4}
\definecolor{red6}{rgb}{1.0,0.5,0.5}
\definecolor{red7}{rgb}{1.0,0.6,0.6}
\definecolor{red8}{rgb}{1.0,0.7,0.7}
\definecolor{red9}{rgb}{1.0,0.8,0.7}
\definecolor{red10}{rgb}{1.0,0.9,0.9}
\definecolor{blue1}{rgb}{0.00,0.0,1.0}
\definecolor{blue2}{rgb}{0.05,0.15,1.0}
\definecolor{blue3}{rgb}{0.15,0.30,1.0}
\definecolor{blue4}{rgb}{0.15,0.45,1.0}
\definecolor{blue5}{rgb}{0.20,0.60,1.0}
\definecolor{blue6}{rgb}{0.25,0.75,1.0}
\definecolor{blue7}{rgb}{0.30,0.90,1.0}
\definecolor{blue8}{rgb}{0.35,0.93,1.0}
\definecolor{blue9}{rgb}{0.40,0.96,1.0}
\definecolor{blue10}{rgb}{0.45,0.99,1.0}

\begin{frontmatter}
\maketitle

\begin{abstract}
\small{
Motivated by the seemingly high accuracy levels of machine learning models in Moldavian versus Romanian dialect identification and the increasing research interest on this topic, we provide a follow-up on the Moldavian versus Romanian Cross-Dialect Topic Identification (MRC) shared task of the VarDial 2019 Evaluation Campaign. The shared task included two sub-task types: one that consisted in discriminating between the Moldavian and Romanian dialects and one that consisted in classifying documents by topic across the two dialects of Romanian. Participants achieved impressive scores, e.g.~the top model for Moldavian versus Romanian dialect identification obtained a macro $F_1$ score of 0.895. We conduct a subjective evaluation by human annotators, showing that humans attain much lower accuracy rates compared to machine learning (ML) models. Hence, it remains unclear why the methods proposed by participants attain such high accuracy rates. Our goal is to understand $(i)$ why the proposed methods work so well (by visualizing the discriminative features) and $(ii)$ to what extent these methods can keep their high accuracy levels, e.g.~when we shorten the text samples to single sentences or when we use tweets at inference time. A secondary goal of our work is to propose an improved ML model using ensemble learning. Our experiments show that ML models can accurately identify the dialects, even at the sentence level and across different domains (news articles versus tweets). We also analyze the most discriminative features of the best performing models, providing some explanations behind the decisions taken by these models. Interestingly, we learn new dialectal patterns previously unknown to us or to our human annotators. Furthermore, we conduct experiments showing that the machine learning performance on the MRC shared task can be improved through an ensemble based on stacking.}

\end{abstract}
\end{frontmatter}

\section{Introduction}

In recent years, we have witnessed an increasing interest in spoken or written dialect identification, proven by a high number of evaluation campaigns \cite{Ali-ASRU-2017,Malmasi-VarDial-2016,Rangel-CLEF-2017,Zampieri-VarDial-2017,Zampieri-VarDial-2018,Zampieri-VarDial-2019,Gaman-VarDial-2020,Chakravarthi-VarDial-2021} with more and more participants. In this paper, we explore the Moldavian versus Romanian Cross-Dialect Topic Identification (MRC) shared task, which was introduced as a task in the VarDial 2019 evaluation campaign \cite{Zampieri-VarDial-2019}, following the release of the MOROCO data set \cite{Butnaru-ACL-2019}. The shared task included two sub-task types: one that consisted in discriminating between the Moldavian and the Romanian sub-dialects and one that consisted in classifying documents by topic across the two sub-dialects of Romanian. However, our primary focus is on the Moldavian versus Romanian dialect identification task, which was further explored in the Romanian Dialect Identification (RDI) shared tasks held at VarDial 2020 \cite{Gaman-VarDial-2020} and VarDial 2021 \cite{Chakravarthi-VarDial-2021}.
While MOROCO is a relatively recent data set, the number of works that studied Romanian dialect identification from a computational perspective \cite{Butnaru-ACL-2019,Chifu-VarDial-2019,Onose-VarDial-2019,Tudoreanu-VarDial-2019,Wu-VarDial-2019,Coltekin-VarDial-2020,Popa-VarDial-2020,Rebeja-VarDial-2020,Jauhiainen-VarDial-2020,Zaharia-VarDial-2020,Ceolin-VarDial-2020,Jauhiainen-VarDial-2021,Ceolin-VarDial-2021,Zaharia-VarDial-2021} has constantly grown due to the organization of annual evaluation campaigns on the topic.

Romanian, the language spoken in Romania, belongs to a Balkan-Romance group that emerged in the fifth century \cite{Coteanu-AR-1969}, after it separated from the Western Romance branch of languages. The Balkan-Romance group is formed of four dialects: Aromanian, Daco-Romanian, Istro-Romanian, and Megleno-Romanian. We underline that, within its group, Romanian is referred to as Daco-Romanian. Noting that Moldavian is a sub-dialect of Daco-Romanian, which is spoken in the Republic of Moldova and in northeastern Romania, the Moldavian versus Romanian dialect identification task is actually a sub-dialect identification task. The Moldavian sub-dialect can be delimited from Romanian in large part by its phonetic features, and only marginally, by morphological and lexical features \cite{Pavel-LR-2008}. Hence, it is much easier to distinguish between the spoken Moldavian and Romanian dialects than the written dialects. This is a first hint that discriminating between Moldavian and Romanian is not an easy task, at least from a human point of view. It is important to add the fact that Romania and the Republic of Moldova have the same literary standard \cite{Minahan-R-2013}. In this context, some linguists \cite{Pavel-LR-2008} believe that a dialectal division between the two countries is not justified. However, literary standards in the two countries are continuously evolving and, for example, this has led to different writings of words containing the vocal sound `\^{a}' (in Romanian) or `\^{i}' (in Moldavian) (see Table~\ref{tab_GradCAM_MD}). Moreover, due to the geographical division between the two countries, people often use different words to denote the same concept (see Section~\ref{sec_Discussion}) and it may happen that they do not understand each other when the discussion includes the respective concept. These differences justify a sub-dialectal division between Romanian and Moldavian. Although we often refer to Romanian and Moldavian as dialects to simplify the writing, they are really sub-dialects. Hence, we study the challenging Moldavian versus Romanian written sub-dialect identification task, since the data set available for the experiments is composed of written news articles \cite{Butnaru-ACL-2019}. We naturally assume that the news articles follow the literary standards. Furthermore, named entities are masked in the entire corpus. Considering all these facts, the dialect identification task should be very difficult. We analyze the difficulty of the task from a human perspective by asking human annotators from Romania and the Republic of Moldova to label news articles with the corresponding dialect. Given that the average accuracy of the human annotators is around $53\%$, the human evaluation confirms the difficulty of the task. Interestingly, the machine learning (ML) methods proposed so far \cite{Butnaru-ACL-2019,Chifu-VarDial-2019,Onose-VarDial-2019,Tudoreanu-VarDial-2019,Wu-VarDial-2019,Popa-VarDial-2020,Zaharia-VarDial-2020,Zaharia-VarDial-2021} attain much higher accuracy rates. For example, the top scoring system in the VarDial 2019 evaluation campaign \cite{Tudoreanu-VarDial-2019} obtained a macro $F_1$ score of 0.895 for Moldavian versus Romanian dialect identification. Furthermore, the string kernels baseline proposed by \citet{Butnaru-ACL-2019} seems to perform even better, with a macro $F_1$ score of 0.941. We therefore consider the machine learning systems for Moldavian versus Romanian dialect identification to be unreasonably effective.

We can naturally suppose that the high accuracy rates of the ML systems can be influenced by different factors. The first factor to consider is that the ML models have access to a large training set from which many discriminative features can be learned, including features unrelated to the dialect identification task, such as features specific to the author style. The second factor is that the samples are full-length news articles formed of several sentences. This increases the chance of finding discriminative features in just about every sample. The third factor is that the news articles are collected from different publication sources from Romania and the Republic of Moldova, and an ML system could just learn to discriminate among the publication sources. 

Our main goal is to determine if the machine learning models catch any dialectal clues or if the high accuracy rates are purely based on alternative factors, such as those exemplified above. In order to explain the unreasonable effectiveness of machine learning systems, we conduct a series of comprehensive experiments on MOROCO, considering all the enumerated factors. First of all, we perform experiments considering only the first sentence in each news article, significantly reducing the length of the text samples. Second of all, we test the systems on a new set of tweets from Romania and the Republic of Moldova collected from a different time period, making sure that the publication sources in the training and the test set are different. This generates a cross-domain (or cross-genre) dialect identification task, with the training (source) domain being represented by news articles and the test (target) domain being represented by tweets. Our findings indicate that, even in this difficult cross-domain setting, the ML systems still outperform humans by a significant margin. We therefore delve into analyzing and visualizing the discriminative features of one of the best-performing ML systems. Our analysis indicates that the machine learning models take their decisions mostly based on morphological and lexical features, many of which were previously unknown to us.

Our second goal is to establish if further performance boosts are possible by combining highly accurate models in a single pipeline. To this end, we first reimplement and evaluate most of the top scoring methods from the related literature \cite{Butnaru-ACL-2019,Onose-VarDial-2019,Tudoreanu-VarDial-2019,Wu-VarDial-2019,Popa-VarDial-2020,Zaharia-VarDial-2020,Zaharia-VarDial-2021}. Then, we proceed towards combining the state-of-the-art methods through ensemble learning, considering an ensemble based on plurality voting and an ensemble based on classifier stacking. Our empirical results show that classifier stacking is useful, indicating that the features captured by the various machine learning models, ranging from string kernels to convolutional, recurrent and transformer networks, can complement each other towards making better decisions.

Different from prior works on Romanian dialect identification \cite{Butnaru-ACL-2019,Chifu-VarDial-2019,Onose-VarDial-2019,Tudoreanu-VarDial-2019,Wu-VarDial-2019,Coltekin-VarDial-2020,Popa-VarDial-2020,Rebeja-VarDial-2020,Jauhiainen-VarDial-2020,Zaharia-VarDial-2020,Ceolin-VarDial-2020,Jauhiainen-VarDial-2021,Ceolin-VarDial-2021,Zaharia-VarDial-2021}, we make the following important contributions:
\begin{itemize}
    \item We introduce MOROCO-Tweets, a new set of over 5,000 Moldavian and Romanian tweets, enabling us and future works to study Romanian dialect identification in a cross-genre scenario.
    \item We study the Romanian dialect identification task in new scenarios, considering models trained on sentences (instead of full news articles) and applied on sentences or tweets, showing how performance degrades as the scenario gets more difficult.
    \item We propose an ensemble based on stacked generalization for Romanian dialect identification.
    \item We study how native Romanian or Moldavian speakers compare to the ML models for dialect identification and categorization by topic, showing that there is a significant performance gap in favor of the ML models for dialect identification.
    \item We present Grad-CAM visualizations \cite{Selvaraju-ICCV-2017} revealing dialectal patterns that explain the unreasonable effectiveness of the ML models. The newly discovered patterns were not known to us or to the human annotators.
\end{itemize}

The remainder of this paper is organized as follows. We present related work on dialect identification in Section~\ref{sec_Related_Work}. We describe the machine learning systems and the ensemble learning methods in Section~\ref{sec_Method}. We present the experiments in Section~\ref{sec_Experiments}, followed by a discussion of the most discriminative features in Section~\ref{sec_Discussion}. Finally, we draw our conclusions in Section~\ref{sec_Conclusion}.

\section{Related Work}
\label{sec_Related_Work}

\subsection{Dialect Identification}

Dialect identification has been acknowledged in the computational linguistics community as an important task, with multiple events and shared tasks materializing this acknowledgement \cite{Malmasi-VarDial-2016, Zampieri-VarDial-2017, Zampieri-VarDial-2018, Zampieri-VarDial-2019,Gaman-VarDial-2020, Zampieri-VarDial-2014, Zampieri-LT4VarDial-2015}. Naturally, some of the most wide-spread languages also tend to be the most well-studied in terms of dialect identification from a computational linguistics perspective. 

To our knowledge, it seems that Arabic is one of the most studied languages, considering modern setups, such as social media \cite{Alyami-ICCAIS-2020}, large and diverse corpora, such as QADI \cite{Abdelali-arXiv-2020}, dialect recognition from speech \cite{Hananispoken-NLE-2018, Shon-ICASSP-2020} or dialect identification from travel text and tweets \cite{Mishra-WANLP-2019}. Preliminary works dealing with Arabic dialect identification used various handcrafted and linguistic features. For instance, \citet{Biadsy-CASL-2009} employed a phonotactic approach to differentiate among four Arabic dialects with good accuracy. In the same direction of study, we can also mention the efforts involving experiments on the Arabic Online Commentary Dataset \cite{Zaidan-ACL-HLT-2011, Zaidan-CL-2014}. More recently, \citet{Guellil-CSE-2016} proposed an unsupervised approach for Algerian dialect identification. Another interesting study is conducted by \citet{Salameh-COLING-2018}, where the city of each speaker is identified based on the spoken dialect. The evaluation campaigns \cite{Ali-ASRU-2017, Malmasi-VarDial-2016, Zampieri-VarDial-2017, Zampieri-VarDial-2018, Bouamor-WANLP-2019} represent one more proof that dialect identification is of much interest from the Arabic language perspective, as these campaigns included a shared task for Arabic dialect identification. We note that one of the most successful approaches in the Arabic dialect identification shared tasks is based on string kernels \cite{Butnaru-VarDial-2018,Ionescu-VarDial-2016,Ionescu-VarDial-2017}.

Among the well-studied languages from a dialectal perspective, there is also Chinese. \citet{Tsai-SC-2002} proposed a Gaussian Mixture Bigram Model in the differentiation of three major Chinese dialects spoken in Taiwan. Later, \citet{Ma-ICASSP-2006} had an attempt at distinguishing among three different Chinese dialects from speech. A semi-supervised approach, outperforming the initial Gaussian Mixture Models (GMM) for dialect identification, is introduced by \citet{Mingliang-ICSP-2008}. In \cite{Xia-WCSP-2011}, gender is employed as a factor in deciding the dialect of different Chinese utterances. A more recent work \cite{Jun-AE-2017} employed deep bottleneck features, which are related to the phoneme level. Through deep bottleneck features, an attempt at suppressing the influence of redundant dialect information from features is made by the author.

A number of works targeting dialect identification were also published for Spanish. The first such work \cite{Zissman-ICASSP-1996} aims at differentiating Cuban and Peruvian dialects from Spanish. The same task is addressed later by \citet{Torres-ODYSSEY-2004}, with an approach based on GMMs, however less accurate than that of \citet{Zissman-ICASSP-1996}. In \citet{Huang-INTERSPEECH-2006}, GMMs with mixture and frame selection are used for Latin-American Spanish dialect identification. More recently, \citet{Francom-LREC-2014} introduced the ACTIV-ES corpus, with informal language records of Spanish speakers from Argentina, Mexico and Spain.

MOROCO \cite{Butnaru-ACL-2019}, the data set on which the current study is based on, comes as a response to the increasing interest in dialect identification with many research efforts for languages such as Arabic \cite{Zaidan-ACL-HLT-2011,Alsarsour-LREC-2018, Bouamor-LREC-2018}, Spanish \cite{Francom-LREC-2014}, Indian \cite{Kumar-WILDRE4-2018} and Swiss \cite{Samardzic-LREC-2016}, trying to attract interest towards under-studied languages such as Romanian.

\subsection{Romanian Sub-Dialect Identification}

The classification of Romanian in four dialects, i.e.~Daco-Romanian, Macedo-Romanian, Aromanian and Megleno-Romanian, has been studied from a purely linguistic perspective for a few decades \cite{Caragiu-ESE-1975, Petrovici-EA-1970, Puscariu-M-1976}. In a modern linguistic work \cite{Lozovanu-IJSSH-2012} that studied Romanian and its dialects, the authors addressed the subject from a geographical, historical and etymological angle. In another modern study, \citet{Nisioi-LaTeCH-2014} proposed a quantitative approach in the investigation of the syllabic structure of the Aromanian dialect, proposing a rule-based algorithm for automatic syllabification. The aforementioned works are valuable studies performed from a social sciences perspective. However, we are interested in the computational nature of differentiating among Romanian and its dialects or sub-dialects. In this regard, to our knowledge, there is one single work \cite{Ciobanu-LREC-2016} to study Romanian dialects from a computational linguistics perspective, before the VarDial 2019 evaluation campaign~\cite{Zampieri-VarDial-2019}. \citet{Ciobanu-LREC-2016} offer a comparative analysis of the phonetic and orthographic discrepancies between various Romanian dialects. However, the data set used in their endeavour to automatically differentiate among the aforementioned dialects, is rather small, containing only 108 words.

\citet{Butnaru-ACL-2019} introduced MOROCO, a data set of 33,564 online news reports collected from Romania and the Republic of Moldova. For each news article, the data set provides dialect labels as well as category labels. 
The authors applied two effective approaches in tackling the problems of dialect identification and categorization by topic: a character-level convolutional neural network, inspired by \citet{Zhang-NIPS-2015}, and a simple Kernel Ridge Regression with custom string kernels, following \citet{Popescu-BEA8-2013}. We note that the data set proposed by \citet{Butnaru-ACL-2019} was also used as benchmark in the first shared task on Moldavian versus Romanian Cross-Dialect Topic Identification (MRC), generating an additional set of publications \cite{Chifu-VarDial-2019, Onose-VarDial-2019, Tudoreanu-VarDial-2019, Wu-VarDial-2019}. In the MRC shared task, the following sub-tasks were proposed: binary classification by dialect and cross-dialect categorization by topic. The participants proposed various approaches for the MRC shared task, ranging from various deep learning models based on word embeddings \cite{Onose-VarDial-2019} or character embeddings \cite{Tudoreanu-VarDial-2019} to shallow Support Vector Machines based on character n-grams \cite{Wu-VarDial-2019} and voting schemes based on a set of handcrafted statistical features \cite{Chifu-VarDial-2019}. 

The set of tweets collected for this work was used as test set in the Romanian Dialect Identification (RDI) task organized at VarDial 2020 \cite{Gaman-VarDial-2020}. The original training and validation sets in MOROCO \cite{Butnaru-ACL-2019} were used as training set, whereas for the validation, we provided both the test set in the original MOROCO split as well as 200 tweets from MOROCO-Tweets. Our logic was that the provided in-domain data will help participants to achieve better performance in the final evaluation round.
Among the interesting submissions received for RDI 2020, we acknowledge the SVM model of \citet{Popa-VarDial-2020}, which combines the powers of three multilingual transformer models trained on Romanian text samples (i.e.~BERT, XLM and XLM-R) and two monolingual models (cased and uncased Romanian BERT) that targeted only Romanian during training. Furthermore, the authors used each sentence in the examples provided for training as standalone samples and they also employ decision thresholds at prediction time, aiming to maximize the macro $F_1$ score. The architectural choices and preprocessing, placed \citet{Popa-VarDial-2020} second in the RDI track organized at VarDial 2020. Another ensemble that participated in the 2020 RDI task was composed of two TF-IDF encoders and a five-layer neural network \cite{Rebeja-VarDial-2020}. With separate encoders for Romanian and Moldavian, the text peculiarities in each of the two dialects are independently captured. The results, however, show a rather poor ability of the chosen model to discriminate among Romanian and Moldavian sentences. One explanation for the generalization issues of the model is in the lack of strong textual markers to differentiate, in writing, among the two dialects of interest. In the same spin of the RDI task, \citet{Jauhiainen-VarDial-2020} proposed a method that relies on the product of relative frequencies of character n-grams. Their approach achieves an $F_1$ score that is $10\%$ higher than the one obtained by \citet{Rebeja-VarDial-2020} and more than $10\%$ below the $F_1$ score obtained by \citet{Popa-VarDial-2020}. A different approach is proposed by \citet{Zaharia-VarDial-2020}, who employ features ranging from character embeddings to Fast Text word embeddings \cite{Bojanowski-TACL-2017} and transformer embeddings obtained through the fine-tuning of Romanian BERT \cite{Dumitrescu-EMNLP-2020}. All these complementary types of embeddings are then fed into a Bidirectional LSTM network suited for the classification by dialect of the provided samples. Perhaps surprisingly, the Na\"{i}ve Bayes model trained on character n-grams presented by \citet{Ceolin-VarDial-2020} achieves a macro $F_1$ score of 0.667, surpassing most of the previously described solutions that rely, to a certain degree, on deep learning or at least on more complex techniques.

At VarDial 2021 \cite{Chakravarthi-VarDial-2021}, the RDI task was reiterated for the third time, with more training data consisting of the entire MOROCO data set \cite{Butnaru-ACL-2019}. The set of tweets collected for the current work was provided as validation data, while for the final testing, we collected a new set of tweets. Our intuition was that providing participants with more tweets for validation, that could also be used for training, will lead to an important performance boost. However, compared to the overall results obtained at the RDI task of VarDial 2020, the results did not improve by a significant margin. The solution submitted at RDI 2021 by \citet{Jauhiainen-VarDial-2021} achieved the best performance, with a macro $F_1$ score of 0.777, which did not fall far from the top scoring systems at RDI 2020. Their approach employed a Na\"{i}ve Bayes model trained using the product of relative frequencies of character n-grams and the language model adaptation method of \citet{Jauhiainen-VarDial-Coling-2018}. Using an approach based on transformers and knowledge distillation, \citet{Zaharia-VarDial-2021} ranked on the second place at RDI 2021, with an $F_1$ score of 0.732. Interestingly, \citet{Ceolin-VarDial-2021} ranked third at the competition, bringing some improvements to the proposed CNN architecture, after using a data augmentation technique consisting of random swaps of the words in each sentence.

In our study, we consider the best performing models in the MRC and RDI shared tasks \cite{Onose-VarDial-2019, Tudoreanu-VarDial-2019, Wu-VarDial-2019,Popa-VarDial-2020,Zaharia-VarDial-2020,Zaharia-VarDial-2021} along with the baselines proposed by \citet{Butnaru-ACL-2019}, combining these approaches into ensemble models based on voting or stacking.

\subsection{Text Classification}

Aside from dialect identification, we are also performing intra and cross-dialect categorization by topic throughout this work. Thus, we consider appropriate to include related work on topic classification.

Text classification is the task of labeling natural language text as pertaining to a predefined number of categories \cite{Sebastiani-ACM-2002,Korde-IJAIA-2012}. As one of the most fundamental tasks in NLP, text classification has been widely studied \cite{Li-arXiv-2020}. Examples of setups and applications are (but not limited to) social media \cite{Chu-ACSAC-2010}, healthcare \cite{Rabhi-MIM-2019,Ubeda-MultilingualBIO-2020,Aji-SMM4H-2021}, information retrieval \cite{Dwivedi-ICTCS-2016}, sentiment analysis \cite{Pang-EMNLP-2002,Liu-MiningTextData-2012,Baid-IJCA-2017,Lin-ICSE-2018,Zhou-CICLing-2020}, content-based recommender systems \cite{Carrillo-TPAAMS-2013}, document summarization \cite{Wu-ICADL-2004,Cao-AAAI-2017}, various business and marketing applications \cite{Engin-ISCIS-2009,Machedon-SocialCom-2013,Halibas-MIC-2018}, legal document categorization \cite{DeAraujo-LREC-2020}. A variety of languages were targeted over time for the popular text classification task, including well-studied languages such as Arabic \cite{Khreisat-DMIN-2006,Elnagar-IPM-2020}, Turkish \cite{Engin-ISCIS-2009,Tufekci-SIU-2012,Kilimci-UBMK-2019,Koksal-INISTA-2020}, French \cite{Rabhi-MIM-2019,Boudjani-ICMLA-2020}, Spanish \cite{Ubeda-MultilingualBIO-2020} and Indian \cite{Swamy-ICICA-2014}, as well as under-resourced languages such as Romanian \cite{Tache-EACL-2021}. The applied classification techniques range from shallow methods, such as Logistic Regression \cite{Hosmer-Wiley-2013}, SVM \cite{Lodhi-JMLR-2002} and Na\"{i}ve Bayes \cite{Sang-IEEE-2006}, to more complex and resource-hungry deep neural networks, such as CNNs \cite{Zhang-NIPS-2015,Chen-NeuralComputAppl-2018}, Hierarchical Attention Networks \cite{Yang-NAACL-2016} and the powerful transformer-based methods that started to dominate the landscape in recent years \cite{Croce-ACL-2020,Chatsiou-arXiv-2020}.

In the resourceful English language, the research community had the means to explore various topic classification techniques, from shallow methods, such as k-nearest neighbors, Multinomial Na\"{i}ve Bayes and decision trees \cite{Rahman-ICASERT-2019}, to deep forests \cite{Daouadi-IS-2021} and Bayesian networks \cite{Nurfikri-ICoICT-2018}. 
Non-English languages are targeted as well for topic classification. For example, advanced and powerful methods such as Hierarchical Attention Networks \cite{Zhou-ICONIP-2017}, transformers \cite{Zhou-ISI-2019} or hybrid Latent Dirichlet Allocation approaches \cite{Hsu-INISTA-2017} are employed in the classification by topic of Chinese text samples. For Spanish, we find a few works focused on topic classification considering both the linguistic approach \cite{Vilares-JIS-2015} as well as the computational alternatives, e.g.~an ensemble of shallow methods \cite{DeLaPena-IberEval-2017}. Categorization by topic has even been explored for understudied languages such as Korean \cite{Suh-2017-JCognSci}, Indonesian \cite{Bakar-IALP-2018} or Romanian \cite{Vasile-NEUREL-2014}, although the number of works are comparably lower. Remarkably, there are a few recent attempts at language-agnostic methods for topic classification \cite{Johnson-WWW-2021,Song-AI-2019}.

We emphasize that in-domain and cross-domain topic classification are common topics in the NLP research community \cite{Glavas-NLPCSS-2017}. Perhaps less common, at least for the Romanian language, the categorization underlined in this work is performed in cross-dialect and intra-dialect setups. To our knowledge, all the other works targeting Romanian cross-dialect and intra-dialect categorization by topic are related to this paper, in that they use the same data set, MOROCO, for training and evaluation \cite{Onose-VarDial-2019,Tudoreanu-VarDial-2019}. 

\section{Methods}
\label{sec_Method}

Throughout this section, we present in detail the most relevant models from the related literature \cite{Butnaru-ACL-2019,Onose-VarDial-2019,Tudoreanu-VarDial-2019,Wu-VarDial-2019,Popa-VarDial-2020,Zaharia-VarDial-2020,Zaharia-VarDial-2021}, which we have selected to build an ensemble. From \citet{Butnaru-ACL-2019}, we select the Kernel Ridge Regression based on string kernels, since this is their best baseline. From \citet{Tudoreanu-VarDial-2019}, the winner of the Moldavian versus Romanian dialect identification sub-task, we select the character-level convolutional neural network (CNN), which is similar in design to the character-level CNN presented by \citet{Butnaru-ACL-2019}. \citet{Onose-VarDial-2019} applied three different deep learning models: a Long Short-Term Memory (LSTM) network, a Bidirectional Gated Recurrent Units (BiGRU) network and a Hierarchical Attention Network (HAN). Since these deep models are quite diverse, we included all their models in our study. From \citet{Wu-VarDial-2019}, we consider the Support Vector Machines based on character n-grams. For efficiency reasons, we employed the dual form of their SVM, which is given by string kernels. Finally, from the more recent approaches proposed at RDI 2020 and 2021 \cite{Popa-VarDial-2020,Zaharia-VarDial-2020,Zaharia-VarDial-2021}, we decided to include a fine-tuned Romanian BERT in our experiments. We did not include other recent models that represent minor variations of the previously selected models to be part of our ensemble. We underline that the considered methods form a broad variety that includes both shallow models based on handcrafted features and deep models based on automatically-learned character or word embeddings. Nevertheless, all methods are essentially based on two steps, data representation and learning, although in some models, e.g.~the character-level CNN, the steps are performed in an end-to-end fashion. We next provide details about the data representations and the learning models considered in our experiments. 

\subsection{Data Representations}
\label{sec_Data_Representations}

\noindent
\textbf{Word Embeddings.} 
Some of the first statistical learning models for building vectorial word representations were introduced in \cite{Bengio-JMLR-2003,Schutze-NIPS-1993}. The goal of vectorial word representations (word embeddings) is to associate similar vectors to semantically related words, allowing us to express semantic relations mathematically in the generated embedding space. After the preliminary work of \citet{Bengio-JMLR-2003} and \citet{Schutze-NIPS-1993}, various improvements have been made to the quality of the embedding and the training time \cite{Collobert-ICML-2008, Mikolov-ICLRW-2013, Mikolov-NIPS-2013, Pennington-EMNLP-2014}, while some efforts have been directed towards learning multiple representations for polysemous words \cite{Huang-ACL-2012, Reisinger-NAACL-2010, Tian-COLING-2014}. These improvements, and many others not mentioned here, have been extensively used in various NLP tasks \cite{Butnaru-Access-2019, Garg-PNAS-2018, Glorot-ICML-2011, Ionescu-NAACL-2019, Musto-ECIR-2016, Weston-IJCAI-2011, Yang-IRJ-2018}.

In the experiments, we use pre-trained word embeddings as features for the LSTM, BiGRU and HAN models. In the feature extraction step, we employ the same set of distributed word representations as \citet{Onose-VarDial-2019}. We note that these representations are learned from Romanian corpora, such as the corpus for contemporary Romanian language (CoRoLa) \cite{Mititelu-LREC-2018, Pais-PRA-2018}, Common Crawl (CC) and Wikipedia \cite{Grave-LREC-2018}, as well as from data coming from the Universal Dependencies project \cite{Nivre-LREC-2016}, that is added to the Nordic Language Processing Laboratory (NLPL) shared repository. In the remainder of this paper, we refer to these representations as CoRoLa, CC and NLPL, respectively.
The distributed representation of the words in CoRoLa was learned using a feed-forward neural network, with words being initially represented as sums of the character n-grams \cite{Bojanowski-TACL-2017} in each word \cite{Pais-PRA-2018}. 
The CoRoLa embeddings used in this work have an embedding size of 300, with a vocabulary size of 250,942 tokens. With the same embedding size, but with a vocabulary that is almost 10 times larger (i.e.~2 millions of words), CC \cite{Grave-LREC-2018} is the second set of pre-trained Romanian word vectors that we have tried out. As we have previously mentioned, CC has been trained on Common Crawl and Wikipedia, using FastText \cite{Bojanowski-TACL-2017, Joulin-EACL-2017}. The third set of word embeddings used in our experiments comes from the NLPL repository and contains vectors of size 100, with the biggest vocabulary of the three, i.e.~2,153,518 words. The NLPL embeddings are trained on the Romanian CoNLL17 corpus \cite{Zeman-CONLL-2018}, using   the Skip-gram model from \emph{word2vec} as learning technique \cite{Mikolov-ICLRW-2013}.

\noindent
\textbf{Character Embeddings.} Some of the pioneering works in language modeling at the character level are \cite{Gasthaus-DCC-2010, Wood-ICML-2009}. To date, characters proved useful in a variety of neural models, such as Recurrent Neural Networks (RNNs) \cite{Sutskever-ICML-2011}, LSTM networks \cite{Ballesteros-EMNLP-2015, Ling-EMNLP-2015}, CNNs \cite{Zhang-NIPS-2015,Kim-AAAI-2016} and transformer models \cite{Al-Rfou-AAAI-2019}. Characters are the smallest units necessary in building words that exist in the vocabulary, regardless of language, as the alphabet changes only slightly across many languages. Thus, knowledge of words, semantic structure or syntax is not required when working with characters. Robustness to spelling errors and words that are outside the vocabulary \cite{Ballesteros-EMNLP-2015} constitute other advantages explaining the growing interest for using characters as features. 

In our paper, we employ three models working at the character level, an SVM and a KRR based on character n-grams \cite{Wu-VarDial-2019}, as well as a character-level CNN \cite{Butnaru-ACL-2019,Tudoreanu-VarDial-2019}. The CNN is equipped with a character embedding layer, generating a 2D representation of text that is further processed by the convolutional layers. We provide additional details about the CNN in Section~\ref{sec_Learning_Models}.

\noindent
\textbf{String Kernels.} \citet{Lodhi-NIPS-2001, Lodhi-JMLR-2002} introduced string kernels as a means of comparing two documents, based on the inner product generated by all substrings of length $n$, typically known as n-grams. Of interest in determining the similarity are the n-grams that the two documents have in common. The authors applied string kernels in a text classification task with promising results. Since then, string kernels have found many applications, from protein classification \cite{Zaki-AB-2005} and learning semantic parsers \cite{Kate-ACL-2006} to tasks as complex as recognizing famous pianists by their playing style \cite{Saunders-ECML-2004} or dynamic scene understanding \cite{Brun-TCSVT-2014}. Other applications of the method include various NLP tasks across different languages, e.g.~sentiment analysis \cite{Gimenez-EACL-2017, Ionescu-EMNLP-2018, Popescu-KES-2017}, authorship identification \cite{Sanderson-EMNLP-2006}, automated essay scoring \cite{Cozma-ACL-2018}, sentence selection \cite{Masala-KES-2017}, native language identification \cite{Popescu-BEA8-2013,Ionescu-EMNLP-2014,Ionescu-COLI-2016,Ionescu-BEA-2017} and dialect identification \cite{Butnaru-ACL-2019,Butnaru-VarDial-2018,Ionescu-VarDial-2017}.
Many improvements have also been added, incrementally, to the original method. These target the space usage \cite{Belazzougui-A-2017}, versatility \cite{Elzinga-TCS-2013} and time complexity \cite{Popescu-KES-2017,Singh-ECML-2017}. 

In this work, we employ string kernels as described in \cite{Butnaru-ACL-2019}, specifically using the efficient algorithm for building string kernels of \citet{Popescu-KES-2017}. We emphasize that the number of character n-grams is usually much higher than the number of samples, so representing the text samples as feature vectors may require a lot of space. String kernels provide an efficient way to avoid storing and using the feature vectors (primal form), by representing the data though a kernel matrix (dual form). Each cell in the kernel matrix represents the similarity between some text samples $x_i$ and $x_j$. In our experiments, we compute the similarity as the presence bits string kernel \cite{Popescu-BEA8-2013}. For two strings $x_i$ and $x_j$ over a set of characters $S$, the presence bits string kernel is defined as follows:
\begin{equation}\label{eq_str_kernel_presence}
k^{0/1}(x_i, x_j)=\sum\limits_{g \in S^n} \mbox{\#}(x_i, g) \cdot \mbox{\#}(x_j, g),
\end{equation}
where $n$ is the length of n-grams and $\mbox{\#}(x,g)$ is a function that returns 1 when the number of occurrences of n-gram $g$ in $x$ is greater than 1, and 0 otherwise.

\subsection{Learning Models}
\label{sec_Learning_Models}

While there is a broad spectrum of machine learning models, e.g.~\cite{Hussein-IJCSM-2020,Shamis-IJCSM-2020}, we only consider models that have been previously used with success for Romanian dialect identification. Additionally, we integrate the individual models presented below into ensembles.

\noindent
\textbf{Support Vector Machines.} The objective of Support Vector Machines (SVM) is to find a hyperplane that best classifies the data points provided in the training phase into two classes \cite{Cortes-ML-1995}. To ensure a good generalization capability, the SVM aims at maximizing the margin that separates the points in both classes. The margin is chosen based on the points that are closest to the decision boundary. These points are called \emph{support vectors}, and, not only do they give the name of the method, but they also influence the orientation and position of the hyperplane that is eventually used for classification during inference. Through the kernel trick, the SVM gains the power to classify data that is not linearly separable, since the data is mapped into a higher-dimensional space, where it becomes separable using a hyperplane \cite{Taylor-CUP-2004}. For multi-class classification, multiple SVM classifiers need to be trained in a one-versus-one or one-versus-rest scheme. In our text categorization by topic experiments, we employ the one-versus-one scheme. Instead of using a standard kernel, we employ the SVM with the custom string kernel based on character n-grams defined in Equation~\ref{eq_str_kernel_presence}. We note that our dual SVM based on string kernels is mathematically equivalent to the primal SVM based on character n-grams employed by~\citet{Wu-VarDial-2019}. We prefer the dual SVM because it is more computationally efficient, as explained in detail by~\citet{Ionescu-COLI-2016}.


\noindent
\textbf{Kernel Ridge Regression.} Ridge Regression \cite{Hoerl-Technometrics-1970}, or linear regression with $L_2$ regularization for overfitting prevention, has been combined with the kernel trick \cite{Saunders-ICML-1998}, enabling the method to capture non-linear relations between features and responses. The kernel version, known as Kernel Ridge Regression (KRR), is a state-of-the-art technique \cite{Taylor-CUP-2004} used in several recent works \cite{Butnaru-ACL-2019,Ionescu-EMNLP-2018,Ionescu-COLI-2016} with very good results. KRR can be seen as a generalization of simple Ridge Regression, learning a function in the Hilbert space described by the kernel. The learned function is either linear or non-linear, with respect to the original space, depending on the considered kernel \cite{Taylor-CUP-2004}.
Although KRR can be used with any kernel function, we employ the KRR based on the kernel defined in Equation~\ref{eq_str_kernel_presence}, as previously proposed by \citet{Butnaru-ACL-2019}. In order to repurpose the trained regressor as a (binary) classifier, we round the predicted continuous values to the values in the set $\{-1, 1\}$. For the multi-class text categorization by topic tasks, we employ KRR in a one-versus-rest scheme.

\noindent
\textbf{Convolutional Neural Networks.} A type of artificial neural network based on convolving multiple sets of filters in a sequential manner is represented by the convolutional neural network. The rectified outputs yielded by the convolution operation are called activation maps and they are subject to pooling operations, which provide a downscaled version of the activation maps, implicitly reducing the amount of parameters and computations further used in the network. After repeating a number of convolutional blocks consisting of convolutions and pooling operations, a sequence of fully-connected layers typically follows, with the last layer having a number of units equal to the number of classes in the data set. Because CNNs are inspired by the mammalian visual cortex \cite{Bengio-FTML-2009,Fukushima-BC-1980}, such models have been found suitable, initially, for image classification \cite{Krizhevsky-NIPS-2012, Lawrence-TNN-1997, LeCun-NC-1989, LeCun-CVPR-2004}. Afterwards, this approach has been adapted for natural language processing (NLP) problems \cite{Zhang-NIPS-2015,Kim-EMNLP-2014}. In NLP, the meaning of the inputs changes: instead of image pixels, we have documents represented as a matrix, using either word \cite{DosSantos-COLING-2014} or character embeddings \cite{Zhang-NIPS-2015}.

One of the models that we employ in the experiments is a character-level CNN \cite{Zhang-NIPS-2015} with squeeze-and-excitation (SE) blocks, introduced by \citet{Butnaru-ACL-2019}. Our motivation for this choice of algorithm lies in $(i)$ the good results obtained on MOROCO by \citet{Butnaru-ACL-2019} and by \citet{Tudoreanu-VarDial-2019}, and also, in $(ii)$ the interpretability of the model through visualization techniques. We used the latter feature to get a better understanding of the CNN model's effectiveness in Section \ref{sec_Discussion}, based on Grad-CAM visualizations \cite{Selvaraju-ICCV-2017}.

\noindent
\textbf{Long Short-Term Memory Networks.} Recurrent Neural Networks (RNNs) \cite{Werbos-NN-1988} represent a type of neural model that operates at the sequence level, achieving state-of-the-art performance on language modeling tasks \cite{Chung-DLRL-2014,Weiss-ACL-2018}, among other problems involving time series. Their effectiveness is constrained by the length of the input sequence. RNNs must use context in order to make predictions, while they also need to learn the context itself, which can lead to vanishing gradients problems \cite{Hochreiter-FGDRNN-2001}, a major drawback of simple RNNs. This is solved in Long Short-Term Memory networks (LSTMs) \cite{Hochreiter-NC-1997}, which rely on an RNN architecture that uses a more complex structure for its base units. An LSTM unit has a cell that acts as a memory element, remembering dependencies in the input. The amount of information stored in this cell and its overall impact is controlled through three gates acting as regulators. The input and output gates control and select the information to be added into and outside of the cell. Later versions of LSTMs also use forget gates, enabling the cell to reset its state for optimization reasons \cite{Gers-NC-2000, Greff-TNNLS-2016}. With these modifications in terms of structure and computation, LSTMs are able to selectively capture long-term dependencies without the technical challenges faced when working with simple RNNs, i.e. exploding and vanishing gradients. \citet{Onose-VarDial-2019} showed that LSTMs are also useful in the dialect identification and categorization sub-tasks on the MOROCO data set. Hence, using this type of network in our experiments has been inspired by \citet{Onose-VarDial-2019}.

\noindent
\textbf{Bidirectional Gated Recurrent Units.} Gated Recurrent Units (GRUs) \cite{Cho-EMNLP-2014} implement a simplified version of LSTMs having only input and forget gates, i.e.~the output gate is excluded. With fewer parameters than LSTMs, the performance achieved by GRUs on various tasks, e.g.~speech recognition, is similar to the one achieved by LSTMs \cite{Ravanelli-TETCI-2018}. Moreover, GRUs tend to outperform LSTMs on small data sets \cite{Chung-DLRL-2014}. The roles seem reversed for problems such as language recognition \cite{Weiss-ACL-2018} or neural machine translation \cite{Britz-EMNLP-2017}. We note that GRUs, as well as other types of RNNs, can use a bidirectional architecture, an adjustment made with the aim of addressing the need of knowing both the previous and the next context to understand the current word. Thus, a bidirectional Gated Recurrent Unit (BiGRU) model is composed of two vanilla GRUs, one with forward activations (i.e.~getting information from the past) and one with backward activations (i.e.~getting information from the future) \cite{Nussbaum-INTERSPEECH-2016}. BiGRUs are among the models that proved their efficiency in the experiments conducted by \citet{Onose-VarDial-2019} on MOROCO, which is why we decided to include the BiGRU architecture in our set of models.

\noindent
\textbf{Hierarchical Attention Networks.} Proposed by \citet{Yang-NAACL-2016}, Hierarchical Attention Networks (HANs) have been initially applied in document classification. The success obtained on this task is explained by the natural approach taken in HANs, reflecting the structure of documents through attention mechanisms applied at two levels: for words that form sentences and for sentences as components of documents. In the case of HANs, the attention mechanism uses context to spot relevant sequences of tokens in a given sentence or document. Essentially, the same algorithms, namely encoding and selection by relevance, are applied twice, at the word level and also at the sentence level \cite{Yang-NAACL-2016}. As for the previously described methods, i.e.~LSTM and BiGRU, the inclusion of HAN in our set of models to be used in the experiments has its motivation in the results obtained by \citet{Onose-VarDial-2019}.

\noindent
\textbf{Romanian BERT.} The transformer architecture was introduced by \citet{Vaswani-NIPS-2017} and showed a remarkable boost in performance compared to the state-of-the-art at the moment. One year later, \citet{Devlin-NAACL-2018} applied the bidirectional training of transformers to address language modeling. They named the new model BERT (Bidirectional Encoder Representations from Transformers), and ever since, BERT was adopted by many NLP researchers as a state-of-the-art transformer-based model. Perhaps one of the most beneficial features of BERT is represented by its multilingual training setup, comprising more than 100 languages. 
In recent years, more and more monolingual flavours of BERT started to be released, e.g.~BERTje for Dutch \cite{DeVries-arXiv-2019}, CamemBERT \cite{Martin-ACL-2019} and FlauBERT for French \cite{Le-LREC-2019}, AlBERTo for Italian \cite{Polignano-CLICIT-2019}, among others. Of particular interest to our work is the Romanian adaption of BERT \cite{Dumitrescu-EMNLP-2020}, which was trained on more than 15GB worth of Romanian data and has been shown to surpass its multilingual counterpart in many tasks, e.g.~named entity recognition \cite{Dumitrescu-EMNLP-2020}. In this work, we fine-tune the Romanian BERT (Ro-BERT) to be able to distinguish among the Romanian and Moldavian dialects. Another classification setup for which we fine-tune Ro-BERT is categorization by topic, where the model learns to discriminate among the six categories available in the data set. In order to obtain the probability for each class, we append a Softmax layer to BERT, either with 2 neurons for dialect identification or 6 neurons for classification by topic.

\begin{figure}
\centering
\includegraphics[width=0.9\linewidth]{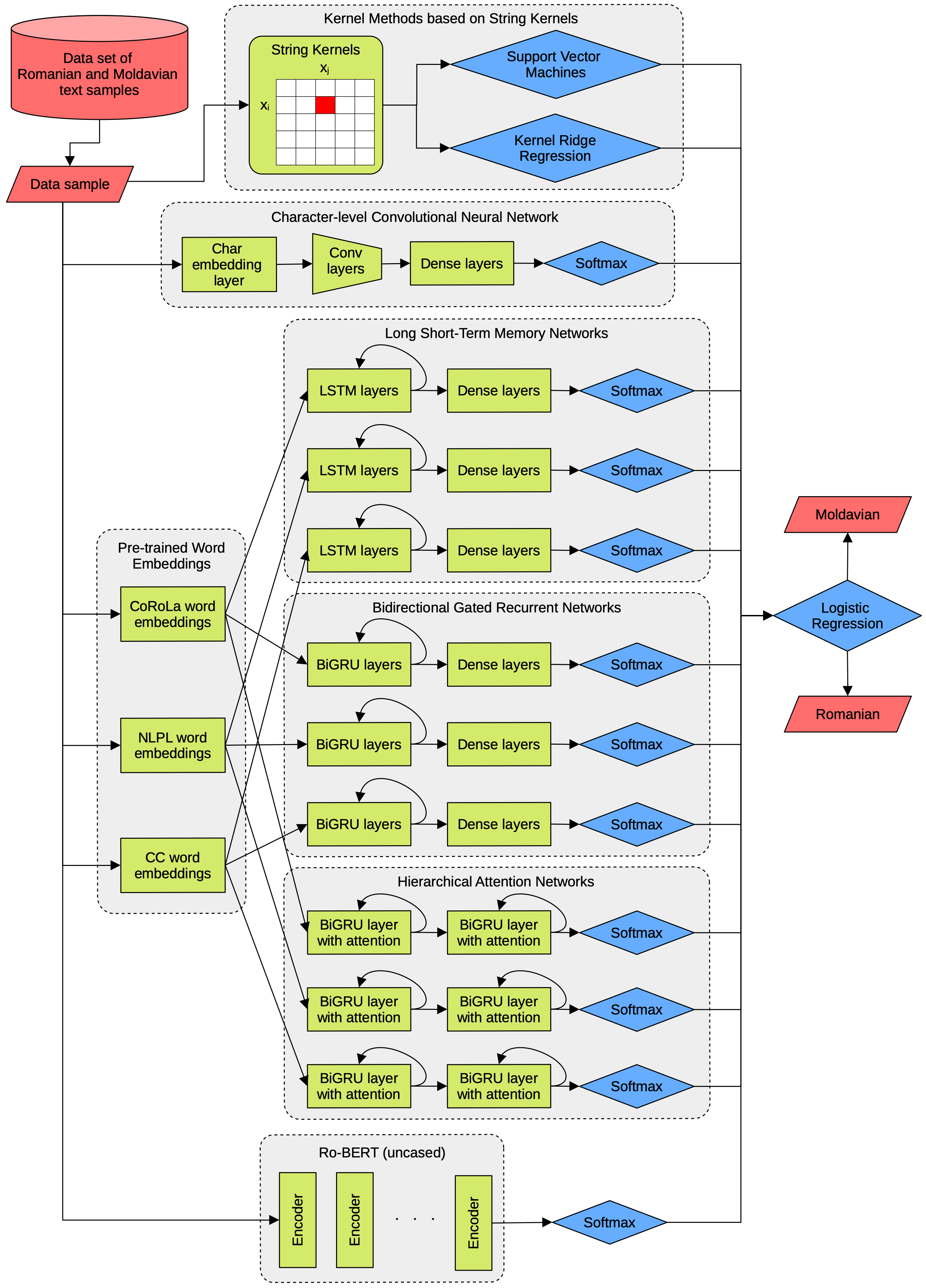}
\caption{Overview of the proposed pipeline based on stacked generalization. Although the pipeline is illustrated for the task of dialect identification, the same architecture is trained for in-domain and cross-domain categorization by topic. Best viewed in color.}
\label{fig_pipeline}
\end{figure}

\noindent
\textbf{Ensemble Models.} The main idea behind ensemble models is to combine multiple learning techniques in order to obtain a model that achieves better results than any of its individual components \cite{Opitz-JAIR-1999, Rokach-AIR-2010}. The model obtained via ensemble learning is typically more stable and robust \cite{Gashler-ICMLA-2008}. There is proof that a significant diversity among the component models of an ensemble leads to better results than in the case where similar techniques are brought together into an ensemble \cite{Kuncheva-ML-2003,Sollich-NIPS-1996}. We use this hypothesis in the experiments conducted in this work. More precisely, our models cover different features as input, from the basic character-level properties of string kernels to the hierarchical selection of words and sentences of HAN. Furthermore, not only that we employ a diversity in the types of features, but we also use different, complementary learning techniques, ranging from shallow models, such as SVM and KRR, to deep models, such as CNN, RNN and BERT. 

We underline that our first motivation for including the above models into our ensemble is that diversity is more likely to generate an ensemble that surpasses its components in terms of accuracy. Our second motivation for including the specified models is that these have been used in top ranking systems at the MRC shared task. Hence, the ensemble is more likely to achieve state-of-the-art performance.

Plurality voting is one of the ensemble approaches that we choose for our experiments. In this approach, the models in the ensemble simply vote with equal weights. The second ensemble learning approach that we consider for the experiments is stacking. Stacked generalization or stacking is an ensemble learning method that learns how to blend the predictions provided by multiple models, through meta-learning \cite{Wolpert-NN-1992}. In our case, the predictions from all the models presented above (SVM, KRR, CNN, LSTM, BiGRU, HAN and Ro-BERT) are taken into consideration. These are known as level-zero models. We underline that we use three types of pre-trained word embeddings for LSTM, BiGRU and HAN, generating a total of nine recurrent models. We note that stacking is different from bagging in that the machine learning models forming the ensemble are different from each other, while being trained on the same data. Stacking uses a meta-model (also known as level-one model) to harness the capabilities of the level-zero models. We employ Multinomial Logistic Regression as our meta-classifier. As input to the meta-model, we consider a vector containing the hard labels as well as the soft scores (class probabilities) provided by each model. Through the meta-model, stacking can learn when to use or trust each model in the ensemble, thus being able to make predictions having superior performance than any single model in the ensemble. Our ensemble learning pipeline based on stacking is illustrated in Figure~\ref{fig_pipeline}.

We emphasize that stacking is suitable when there are multiple distinct models achieving good performance levels, but on different data samples. In other words, if predictions of the level-zero models have a low correlation, then it is likely to achieve superior results through stacking. Since our level-zero models are different from each other, we believe that stacked generalization is a suitable method to obtain a good ensemble. Although stacking is designed to increase performance, to our knowledge, there is no guarantee that it will lead to superior results in all cases. Finally, we underline that ensemble learning has not been studied on MOROCO before our work. Hence, this is the first study to test out the effectiveness of ensemble learning in Romanian dialect identification.

\section{Experiments}
\label{sec_Experiments}

\subsection{Data Sets}

The Moldavian and Romanian Dialectal Corpus (MOROCO)\footnote{https://github.com/butnaruandrei/MOROCO} \cite{Butnaru-ACL-2019} is the main data set employed in the experiments conducted in this work. The corpus was collected from the top five news websites from Romania and the Republic of Moldova as data sources, using each country's web domain (\emph{.ro} or \emph{.md}) to automatically label the news articles by dialect. \citet{Butnaru-ACL-2019} also provide topic labels, assigning each news article in the corpus to one of six categories: \textit{culture}, \textit{finance}, \textit{politics}, \textit{science}, \textit{sports}, \textit{tech}. A minimum of approximately 2000 samples per dialect is obtained for each topic. The corpus was automatically pre-processed to remove named entities. MOROCO is comprised of 33,564 news articles, with an official split of 21,719 training samples, 5921 validation samples and 5924 test samples.

Although we are primarily interested in the dialect identification task, we present results for the full range of tasks proposed by \citet{Butnaru-ACL-2019}, namely: 
\begin{itemize}
    \item binary discrimination between Romanian (RO) and Moldavian (MD);
    \item Romanian intra-dialect categorization by topic;
    \item Moldavian intra-dialect categorization by topic;
    \item cross-dialect categorization by topic using Moldavian as source and Romanian as target;
    \item cross-dialect categorization by topic using Romanian as source and Moldavian as target.
\end{itemize}

In this paper, we introduce an additional data set composed of tweets collected from Romania and the Republic of Moldova, which allows us to evaluate the machine learning models in a cross-genre dialect identification setting. The tweets were collected from a different time period, helping us to reveal any overfitting behavior of the models. The MOROCO-Tweets\footnote{https://github.com/raduionescu/MOROCO-Tweets} data set is divided into a validation set of 215 tweets and a test set of 5,022 tweets, both having a balanced distribution of Moldavian and Romanian tweets. Indeed, the validation set is composed of 113 Moldavian tweets and 102 Romanian tweets, while the test set is composed of 2,499 Moldavian tweets and 2,523 Romanian tweets. All tweets are pre-processed for named entity removal. We did not collect any topic labels from Twitter, since we are mostly interested in cross-genre dialect identification.

\subsection{Experimental Setup}

We first evaluate the considered machine learning models on MOROCO, using the complete news articles, as in all previous works~\cite{Butnaru-ACL-2019,Onose-VarDial-2019,Tudoreanu-VarDial-2019, Wu-VarDial-2019,Popa-VarDial-2020,Zaharia-VarDial-2020,Zaharia-VarDial-2021}. Since our aim is to determine the extent to which machine learning models attain good performance levels, we consider an additional scenario in which we keep only the first sentence from each news article. This essentially transforms all tasks into sentence-level classification tasks. As we keep the same number of data samples, the sentence-level classification accuracy rates are expected to drop, essentially because there are less patterns in the data. We include an even more difficult evaluation setting, testing the models trained at the sentence-level on tweets, while considering only the dialect identification task. As evaluation metrics, we employ the classification accuracy and the macro $F_1$ score. We note that the macro $F_1$ score is the official metric chosen for the VarDial evaluation campaigns that featured dialect identification tasks using MOROCO as support data set \cite{Zampieri-VarDial-2019,Gaman-VarDial-2020,Chakravarthi-VarDial-2021}.

\subsection{Parameter Tuning}

We have borrowed as many of the hyperparameters as possible from the works~\cite{Butnaru-ACL-2019,Onose-VarDial-2019,Tudoreanu-VarDial-2019,Wu-VarDial-2019,Popa-VarDial-2020,Zaharia-VarDial-2020,Zaharia-VarDial-2021} proposing the models considered in our experiments, trying to replicate the previously reported results as closely as possible. When sufficient details to replicate the results were missing, we tune the corresponding hyperparameters on the validation data. We next present the hyperparameter choices for each machine learning model.

\noindent
\textbf{SVM.} We train an SVM with a pre-computed string kernel with $C = 10^2$, which has been selected via grid search from a range of values starting from $10^{-3}$ to $10^3$, considering a multiplication step of $10$. The string kernel is based on character 6-grams.

\noindent
\textbf{KRR.} For KRR, the only parameter that requires tuning is the regularization $\lambda$. From a set of potential values ranging from $10^{-5}$ to $10^{-1}$, with a multiplication step of $10$, the best $\lambda$ for our setup is $10^{-2}$. As for the SVM, the string kernel used in KRR is based on character 6-grams.


\noindent
\textbf{Character-level CNN.} We employ the same architecture and hyperparameters as \citet{Butnaru-ACL-2019}. For the experiments performed with full news articles, we use an input size of $5000$ characters. For the experiments performed on the first sentence from each news article, we adjust the input size to $1000$ characters. The input layer is followed by an embedding layer, which embeds each character into a vector of $128$ components. The neural architecture consists of three convolutional blocks, each having a convolutional layer with $128$ filters, stride $1$ and filter sizes $7$, $7$ and $3$, respectively. Max pooling with a filter size of $3$ is applied in each convolutional block. After each convolutional block, we insert a Squeeze-and-Excitation block with the reduction ratio set to $r = 64$. Two fully-connected layers follow the convolutional blocks, each having $128$ neural units. Each of these two fully-connected layers is subject to dropout, with the probability of dropping individual units of $0.3$. The final layer in the network is the one used for prediction, having $2$ or $6$ neurons depending on the problem solved, i.e. $2$ neurons for dialect identification and $6$ for categorization by topic. The classification layer is based on Softmax activation. We use a learning rate equal with $2\cdot10^{-4}$ and train the network for $50$ epochs on mini-batches of $128$ samples.

\noindent
\textbf{LSTM, BiGRU and HAN.} The architectures and setup used for these models are identical or very similar in most aspects with those implemented by \citet{Onose-VarDial-2019}. Each model is trained for $20$ epochs with a mini-batch size of $50$ samples. These three models are all bases on word embeddings, and we combine each one of them with the previously presented word embeddings, namely CoRoLa, NLPL and CC, resulting in a set of $9$ independent models.

For the LSTM network, we employ an architecture with two LSTM layers of $256$ and $512$ neurons in this order, both having \emph{tanh} activations. The second LSTM layer is followed by dropout regularization, with a probability of dropping out individual neurons of $0.3$. Two dense layers follow next, the first one having $512$ neural units and ReLU activations. The second fully-connected layer in the architecture is a classification layer. It has Softmax activations and $n=\{2,6\}$ neurons, where $n$ depends on the sub-task for which the network is used, i.e.~$n=2$ for dialect identification and $n=6$ for categorization by topic.

BiGRU consists in a GRU layer of $256$ neurons and a bidirectional GRU layer with $512$ neurons. The activation function used in both layers is \emph{tanh}. Two fully-connected layers of $512$ and $1024$ units are also added to the architecture. For regularization and training acceleration reasons, we apply batch normalization after each GRU layer. We add dropout with a rate of $0.3$ after each dense layer. The last layer relies on Softmax activations, performing the classification task. As for the LSTM, the number of neurons in the last layer is $n=\{2,6\}$.

In HAN, we set the maximum sequence length to $150$ words, which is also valid for the other word-based models described previously, namely LSTM and BiGRU. A sentence encoder using a bidirectional GRU layer with $200$ neural units is employed in the first half of the network. The maximum document size considered for the second half is of $20$ sequences. For the second encoder, i.e.~the document encoder in HAN, we have a similar bidirectional GRU layer, with a size of $200$ units. The prediction layer, which comes in last, has $2$ or $6$ neurons with Softmax activations.

\noindent
\textbf{Ro-BERT.} We fine-tune Romanian BERT with mini-batches of 32 samples each, while allowing a maximum sequence length of 128 tokens. A sample exceeding the maximum sequence length is truncated at the end, whereas a shorter sample is zero-padded until it reaches 128 tokens. During optimization, we employ the Adam optimizer with a learning rate of $5 \cdot 10^{-5}$ and $\epsilon$ equal to $10^{-8}$. We fine-tune the transformers for a maximum of $30$ epochs or until accuracy does not improve for at least 5 epochs.

\noindent
\textbf{Ensemble Models.} While the pluraity voting strategy requires no hyperparameter tuning, the meta-learner used in model stacking, namely Logistic Regression, requires tuning of the regularization parameter $C$ and the penalty. As penalty, we generally obtain better validation results with $L_2$ over $L_1$, except for Moldavian intra-dialect categorization by topic. The parameter $C$ is validated within $10^{-3}$ and $10^3$, considering a step of $10$. Depending on the task, we typically obtain the best validation results with $C=10^{-1}$ or $C=1$. An exceptional case is the sentence-level dialect identification task, where the optimal $C$ is $10^{-3}$.


\subsection{Dialect Identification Results}

\begin{table}[!t]
\begin{center}
\caption{Accuracy rates, macro $F_1$ scores and running times of various machine learning models obtained at test time for the Moldavian versus Romanian dialect identification task. The results are report for three evaluation scenarios: $(i)$ full articles: training and testing on full news articles, $(ii)$ sentences: training and testing on the first sentence from each news article, $(iii)$ tweets: training on the first sentence from each news article and testing on tweets. The best results on each column are highlighted in bold. Training and inference times are measured on a machine with an Intel Xeon E5-2687W v4 3.00 GHz CPU, two Nvidia GeForce GTX 1080Ti GPUs, and 256 GB of RAM.}
\label{tab_results_dialect}
\setlength{\tabcolsep}{3.3pt}
\begin{tabular}{llccccccrr}
\hline
\multicolumn{10}{c}{Dialect Identification} \\
\hline
            &                   & \multicolumn{3}{c}{Accuracy}      & \multicolumn{3}{c}{Macro $F_1$} & \multicolumn{2}{c}{Time} \\
\cline{3-10}
Model       & Embedding         & \multirow{2}{*}{Articles}  & \multirow{2}{*}{Sentences} & \multirow{2}{*}{Tweets}    & \multirow{2}{*}{Articles}  & \multirow{2}{*}{Sentences} & \multirow{2}{*}{Tweets} & \multicolumn{1}{c}{Training} & \multicolumn{1}{c}{Inference}\\
\arrayrulecolor{white}
\cline{1-1}
\arrayrulecolor{gray!40!white}
            &                   &   &  &     &   &  &  & \multicolumn{1}{c}{(total)} & \multicolumn{1}{c}{(per sample)}\\
\hline
SVM         & string kernels    & 0.939     & 0.829     & 0.680     & 0.939     & 0.828     & 0.678     &  152m     &  26ms \\
KRR         & string kernels    & 0.943     & 0.818     & 0.683     & 0.943     & 0.817     & 0.682     &  147m     &  26ms \\
CNN         & characters        & 0.930     & 0.780     & 0.606     & 0.929     & 0.780     & 0.605     &  8m       & 1ms  \\
LSTM        & CoRoLa            & 0.861     & 0.776     & 0.572     & 0.859     & 0.771     & 0.564     &  9m       & 7ms  \\
LSTM        & NLPL              & 0.854     & 0.777     & 0.574     & 0.852     & 0.773     & 0.567     &  10m      & 8ms  \\
LSTM        & CC                & 0.844     & 0.781     & 0.541     & 0.843     & 0.777     & 0.538     &  12m      & 10ms  \\
BiGRU       & CoRoLa            & 0.853     & 0.784     & 0.551     & 0.849     & 0.779     & 0.506     &  12m      & 60ms  \\
BiGRU       & NLPL              & 0.865     & 0.778     & 0.595     & 0.864     & 0.768     & 0.575     &  13m      & 61ms  \\
BiGRU       & CC                & 0.866     & 0.705     & 0.580     & 0.865     & 0.702     & 0.578     &  16m      & 62ms  \\
HAN         & CoRoLa            & 0.702     & 0.705     & 0.527     & 0.697     & 0.701     & 0.524     &  98m      & 97ms  \\
HAN         & NLPL              & 0.696     & 0.695     & 0.510     & 0.694     & 0.692     & 0.495     & 259m      & 100ms  \\
HAN         & CC                & 0.697     & 0.700     & 0.516     & 0.694     & 0.696     & 0.512     & 667m      & 101ms  \\
BERT         & BERT embeddings  & 0.932     & 0.864     & 0.661     & 0.931     & \textbf{0.863}    & 0.656     & 126m      & 5ms  \\
Voting      & All               & 0.937     & 0.810     & 0.571     & 0.936     & 0.800             & 0.534     &  1376m    & 521ms  \\
Stacking    & All               & \textbf{0.946} & \textbf{0.865} & \textbf{0.700} & \textbf{0.946} & \textbf{0.863} & \textbf{0.699}   & 1377m     & 521ms  \\
\hline
\end{tabular}
\end{center}
\end{table}

In Table~\ref{tab_results_dialect}, we present the dialect identification results of various ML methods in three different scenarios. In the first scenario, in which the models are trained and tested on full news articles, there are four individual models that surpass the $90\%$ threshold for both evaluation metrics, namely the SVM, the KRR, the character-level CNN and the fine-tuned Romanian BERT. The ensemble models are also going beyond this threshold. In general, it seems that the dialect identification task on entire news articles is fairly easy. However, the high accuracy rates could also be explained by many other factors, namely by the fact that the models actually discriminate the news articles based on author style, publication source or the discussed subjects, which might be different in the two countries. In order to diminish the effects of such additional factors, we considered two additional scenarios, one that involves training and testing at the sentence level, and one that involves a cross-genre evaluation. In the second scenario, in which the models are trained and tested on sentences, we observe significant performance drops with respect to the first scenario. Indeed, the accuracy rates and the macro $F_1$ scores drop by roughly $10\%$ for almost all models. The only model that does not register such a high performance decrease is HAN, but its scores in the first scenario are quite low. Although it is much harder to recover the author style, the publication source or the subject from the first sentence of each news article, these patterns are not completely eliminated. We therefore consider the third evaluation scenario, in which the models are trained on sentences from MOROCO and tested on tweets collected from different sources and from a different time period. We observe further performance drops in the third scenario. While some models are close to a random chance prediction, e.g.~HAN, other models are close to $70\%$ in terms of both accuracy and macro $F_1$. As shown in Table~\ref{tab_results_human}, the human-level performance in the Moldavian versus Romanian dialect identification task is much under the best performing ML models evaluated on tweets. In order to understand and explain this difference, we analyze the Grad-CAM visualizations \cite{Selvaraju-ICCV-2017} for one of the best performing models, namely the character-level CNN, in Section~\ref{sec_Discussion}. Considering all three evaluation scenarios, the individual models attaining the best results are the SVM and the KRR models, both being based on string kernels. These two models are closely followed by the state-of-the-art Romanian BERT, which even outperforms SVM and KRR at the sentence level. We can say for sure that Ro-BERT precedes the character-level CNN, the latter model being ranked as the fourth best. The plurality voting strategy attains mixed results, failing to surpass the top two individual models in all three evaluation scenarios. However, our ensemble based on stacking seems to be more powerful, achieving the best results in each and every case. 

\subsection{Running Time}

The last two columns in Table~\ref{tab_results_dialect} indicate the training and inference times for the experiments conducted on full news articles. For each method, the reported training time (measured in minutes) represents the total amount of time required to extract features and to learn the model until convergence, while the inference time (measured in milliseconds) represents the average time required to extract features and to predict the label for one news article. All times are measured on machine with an Intel Xeon E5-2687W v4 3.00 GHz CPU, two Nvidia GeForce GTX 1080Ti GPUs, and 256 GB of RAM.

In terms of traning time, the most efficient model is the char-CNN, which is followed by the LSTM and BiGRU models based on various word embeddings. In terms of inference time, the most efficient models are the char-CNN with 1 ms and the BERT model with 5 ms. These models are followed by the LSTMs based on word embeddings, each requiring less than 10 ms, and the SVM and KRR based on string kernels, each requiring 26 ms. The least efficient individual model is clearly HAN. 

We underline that it is natural for the ensembles based on voting and stacking to take more time during training and inference than the individual models forming the ensembles. However, the times computed for the ensembles are slightly below the sum of times computed for the individual models. This happens because some features, namely string kernels, CoRoLa, NLPL and CC, can be extracted only once for all individual models that use these features. The inference time for both ensembles is about 0.5 seconds per news article. We thus believe the ensemble models can be used in practical scenarios without any trouble.

\subsection{Intra-Dialect Categorization Results}

\begin{table}[!t]
\begin{center}
\caption{Accuracy rates and macro $F_1$ scores of various machine learning models obtained at test time for the intra-dialect categorization by topic tasks. The results are report for two evaluation scenarios: $(i)$ full articles: training and testing on full news articles, $(ii)$ sentences: training and testing on the first sentence from each news article. The best results on each column are highlighted in bold.}
\label{tab_results_categorization_intra}
\setlength{\tabcolsep}{5pt}
\begin{tabular}{llccccccccc}
\hline
\multicolumn{10}{c}{Intra-Dialect Categorization by Topic} \\
\hline
            &                   & \multicolumn{4}{c}{Accuracy}              & \multicolumn{4}{c}{Macro $F_1$} \\
\cline{3-10}
Model       & Embedding         & \multicolumn{2}{c}{Full Articles} & \multicolumn{2}{c}{Sentences} & \multicolumn{2}{c}{Full Articles} & \multicolumn{2}{c}{Sentences} \\
\cline{3-10}
            &                   & MD        & RO        & MD        & RO        & MD        & RO        & MD        & RO \\
\hline
\rowcolor{gray!20!white}
SVM         & string kernels    & 0.925     & 0.766     & 0.791     & 0.608     & 0.907     & 0.785     & 0.726     & 0.609 \\
KRR         & string kernels    & 0.923     & 0.743     & 0.821     & 0.591     & 0.900     & 0.782     & 0.769     & 0.610 \\
\rowcolor{gray!20!white}
CNN         & characters        & 0.932     & 0.894     & 0.906     & 0.843     & 0.669     & 0.626     & 0.484     & 0.336 \\
LSTM        & CoRoLa    & \textbf{0.957}& \textbf{0.918}& 0.927     & 0.877     & 0.824     & 0.752     & 0.689     & 0.575 \\
\rowcolor{gray!20!white}
LSTM        & NLPL              & 0.956     & 0.912     & 0.928     & 0.879     & 0.833     & 0.732     & 0.701     & 0.585 \\
LSTM        & CC                & 0.936     & 0.914 & \textbf{0.935}& \textbf{0.880}& 0.690 & 0.745     & 0.745     & 0.602 \\
\rowcolor{gray!20!white}
BiGRU       & CoRoLa            & 0.912     & 0.864     & 0.892     & 0.841     & 0.780     & 0.624     & 0.365     & 0.439 \\
BiGRU       & NLPL              & 0.945     & 0.903     & 0.913     & 0.838     & 0.789     & 0.659     & 0.596     & 0.361 \\
\rowcolor{gray!20!white}
BiGRU       & CC                & 0.926     & 0.821     & 0.846     & 0.825     & 0.698     & 0.306     & 0.434     & 0.422 \\
HAN         & CoRoLa            & 0.873     & 0.809     & 0.870     & 0.843     & 0.476     & 0.381     & 0.471     & 0.363 \\
\rowcolor{gray!20!white}
HAN         & NLPL              & 0.876     & 0.844     & 0.876     & 0.842     & 0.482     & 0.373     & 0.465     & 0.365 \\
HAN         & CC                & 0.873     & 0.826     & 0.867     & 0.826     & 0.484     & 0.399     & 0.464     & 0.398 \\
\rowcolor{gray!20!white}
BERT         & BERT embeddings  & 0.927     & 0.763     & 0.881     & 0.669     &   0.913   & 0.800     & \textbf{0.847}     & 0.678 \\
Voting      & All               & 0.916     & 0.756     & 0.854     & 0.665     & 0.891     & 0.800     & 0.807     & 0.682 \\
\rowcolor{gray!20!white}
Stacking    & All               & 0.928     & 0.783     & 0.881     & 0.668 & \textbf{0.916}& \textbf{0.845} & 0.846 & \textbf{0.689} \\
\hline
\end{tabular}
\end{center}
\end{table}

We report the intra-dialect categorization by topic accuracy rates and macro $F_1$ scores of various models in Table~\ref{tab_results_categorization_intra}. First of all, we note that the models generally attain better results within the Moldavian dialect as opposed to the Romanian dialect. In the first evaluation scenario, which is based on full news articles, all models, except HAN, surpass the $90\%$ threshold in terms of accuracy rate for the Moldavian news articles. On both dialects, the best accuracy rates in the first evaluation scenario are obtained by the LSTM based on CoRoLa embeddings, surpassing even the ensemble models. The LSTM based on CC embeddings attains the top accuracy rates on both dialects in the second evaluation scenario, which is conducted at the sentence level. In general, we observe that deep learning models attain better accuracy rates than the shallow SVM and KRR, while the latter models are ranked second and third after the Romanian BERT model, by macro $F_1$ scores, among all individual models. We underline that the high differences between the classification accuracy, which is equivalent to the micro $F_1$ score, and the macro $F_1$ score of each classifier can be explained by the fact that the topic distribution in MOROCO is unbalanced~\cite{Butnaru-ACL-2019}. The macro $F_1$ score is considered more relevant by the VarDial shared tasks organizers~\cite{Zampieri-VarDial-2019}, as it assigns equal weights to each class. Although SVM and KRR surpass other individual models, the best macro $F_1$ scores in most intra-dialect categorization experiments are attained by the ensemble based on classifier stacking. An exceptional case is the categorization of sentences written in Moldavian, where the macro $F_1$ score obtained by Ro-BERT marginally surpasses the score achieved by our ensemble. Comparing the categorization results of the ML models in the second evaluation scenario with those reported for the human annotators in Table~\ref{tab_results_human}, we observe that the performance gap in favor of the machine learning models is smaller with respect to the gap observed in the dialect identification experiments. We conjecture that this observation indicates that the dialectal features are likely more subtle than the topical features. 

\subsection{Cross-Dialect Categorization Results}

\begin{table}[!t]
\begin{center}
\caption{Accuracy rates and macro $F_1$ scores of various machine learning models obtained at test time for the cross-dialect categorization by topic tasks, namely MD$\rightarrow$RO and RO$\rightarrow$MD. The results are report for two evaluation scenarios: $(i)$ full articles: training and testing on full news articles, $(ii)$ sentences: training and testing on the first sentence from each news article. The best results on each column are highlighted in bold.}
\label{tab_results_categorization_cross}
\setlength{\tabcolsep}{5pt}
\begin{tabular}{llccccccccc}
\hline
\multicolumn{10}{c}{Cross-Dialect Categorization by Topic} \\
\hline
            &                   & \multicolumn{4}{c}{Accuracy}              & \multicolumn{4}{c}{Macro $F_1$} \\
\cline{3-10}
            &                   & \multicolumn{2}{c}{Full Articles} & \multicolumn{2}{c}{Sentences} & \multicolumn{2}{c}{Full Articles} & \multicolumn{2}{c}{Sentences} \\
\cline{3-10}
Model       & Embedding         & MD        & RO        & MD        & RO        & MD        & RO        & MD        & RO \\
\rowcolor{gray!0!white}
            &                   & $\downarrow$ & $\downarrow$ & $\downarrow$ & $\downarrow$ & $\downarrow$ & $\downarrow$ & $\downarrow$ & $\downarrow$ \\
            &                   & RO        & MD        & RO        & MD        & RO        & MD        & RO        & MD \\
\hline
\rowcolor{gray!20!white}
SVM         & string kernels    & 0.659     & 0.811     & 0.485     & 0.638     & 0.647     & 0.739     & 0.467     & 0.561 \\
KRR         & string kernels    & 0.692     & 0.823     & 0.489     & 0.641     & 0.693     & 0.758     & 0.467     & 0.558 \\
\rowcolor{gray!20!white}
CNN         & characters        & 0.856     & 0.905     & 0.832     & 0.859     & 0.461     & 0.589     & 0.317     & 0.418 \\
LSTM        & CoRoLa            & 0.892     & 0.941     & 0.851     & 0.901     & 0.652     & 0.750     & 0.482     & 0.614 \\
\rowcolor{gray!20!white}
LSTM        & NLPL      & \textbf{0.895}& \textbf{0.943}& 0.864     & 0.900     & 0.675     & 0.771     & 0.499     & 0.610 \\
LSTM        & CC                & 0.885     & 0.942     & 0.856 & \textbf{0.907}& 0.594     & 0.755     & 0.469     & 0.617 \\
\rowcolor{gray!20!white}
BiGRU       & CoRoLa            & 0.845     & 0.901     & 0.824     & 0.847     & 0.533     & 0.584     & 0.353     & 0.380 \\
BiGRU       & NLPL              & 0.857     & 0.866     & 0.803     & 0.894     & 0.515     & 0.531     & 0.314     & 0.579 \\
\rowcolor{gray!20!white}
BiGRU       & CC                & 0.842     & 0.833     & 0.789     & 0.817     & 0.376     & 0.531     & 0.318     & 0.579 \\
HAN         & CoRoLa            & 0.780     & 0.801     & 0.795     & 0.823     & 0.284     & 0.311     & 0.281     & 0.333 \\
\rowcolor{gray!20!white}
HAN         & NLPL              & 0.784     & 0.839 & \textbf{0.890}& 0.827     & 0.290     & 0.297     & 0.289     & 0.321 \\
HAN         & CC                & 0.795     & 0.814     & 0.798     & 0.814     & 0.290     & 0.309     & 0.291     & 0.308 \\
\rowcolor{gray!20!white}
BERT      & BERT embeddings     & 0.712     & 0.846     & 0.623     & 0.768     & 0.716     & 0.802     & \textbf{0.607}     & \textbf{0.712} \\
Voting      & All               & 0.709     & 0.844     & 0.567     & 0.725     & 0.709     & 0.781     & 0.544     & 0.666 \\
\rowcolor{gray!20!white}
Stacking    & All               & 0.722     & 0.873     & 0.601     & 0.799 & \textbf{0.727}& \textbf{0.833} & 0.602 & 0.703 \\
\hline
\end{tabular}
\end{center}
\end{table}

In Table~\ref{tab_results_categorization_cross}, we present the accuracy rates and the macro $F_1$ scores of the considered ML models for cross-dialect categorization by topic in two scenarios, one based on full articles and one based on sentences. In general, we notice that most of the patterns observed in the intra-dialect categorization experiments shown in Table~\ref{tab_results_categorization_intra} also apply to the cross-dialect experiments. Indeed, we observe that the deep learning methods typically yield superior accuracy rates with respect to the shallow methods based on string kernels, the best approach in most cases being the LSTM network. Nevertheless, the SVM and the KRR compensate by attaining better macro $F_1$ scores than most of the deep learning models. The two kernel approaches are consistently surpassed by Ro-BERT. As for the in-domain experiments, the ensemble based on stacking yields the top macro $F_1$ scores for both cross-dialect tasks performed on full articles. Ro-BERT slightly surpasses the ensemble meta-learner when it comes to cross-dialect categorization of sentences. In summary, we consider that the idea of combining the models into an ensemble via classifier stacking is very useful. Comparing the cross-dialect categorization results of the ML classifiers at the sentence level with those reported for the human annotators in Table~\ref{tab_results_human}, we emphasize that, at least in terms of the macro $F_1$ metric, humans are generally better.

\subsection{Human Annotation Results}


\begin{table}[!t]
\begin{center}
\caption{Accuracy rates and macro $F_1$ scores of ten human subjects that were asked to annotate 120 sentences with dialectal and categorical labels. The last row indicates the average values computed on all ten annotators.}
\label{tab_results_human}
\begin{tabular}{ccccc}
\hline
\multicolumn{5}{c}{Human Annotated Data} \\
\hline
       & \multicolumn{2}{c}{Accuracy}       & \multicolumn{2}{c}{Macro $F_1$} \\
\cline{2-5}
Annotator ID    & Dialect           & Categorization    & Dialect           & Categorization \\
\rowcolor{gray!0!white}
                & Identification    & by Topic          & Identification    & by Topic \\
\hline
\rowcolor{gray!20!white}
\#A1   & 0.483             & 0.708             & 0.482             & 0.707 \\
\rowcolor{gray!0!white}
\#A2   & 0.533             & 0.700             & 0.527             & 0.697 \\
\rowcolor{gray!20!white}
\#A3   & 0.583             & 0.742             & 0.582             & 0.736 \\
\rowcolor{gray!0!white}
\#A4   & 0.483             & 0.717             & 0.481             & 0.717 \\
\rowcolor{gray!20!white}
\#A5   & 0.550             & 0.683             & 0.436             & 0.681 \\
\rowcolor{gray!0!white}
\#A6   & 0.567             & 0.633             & 0.559             & 0.627 \\
\rowcolor{gray!20!white}
\#A7   & 0.550             & 0.750             & 0.549             & 0.748 \\
\rowcolor{gray!0!white}
\#A8   & 0.500             & 0.667             & 0.499             & 0.669 \\
\rowcolor{gray!20!white}
\#A9   & 0.542             & 0.717             & 0.499             & 0.717 \\
\rowcolor{gray!0!white}
\#A10   & 0.525             & 0.725             & 0.521             & 0.722 \\
\rowcolor{gray!20!white}
Average         & 0.532             & 0.704             & 0.513             & 0.702 \\
\hline
\end{tabular}
\end{center}
\end{table}

We have asked ten human subjects to manually annotate a subset of 120 randomly selected samples from the MOROCO data set. Among the subjects involved in the annotation task, there were nine native speakers of Romanian and one native speaker of Moldavian. All annotators understood the task and the presented examples in both dialects. The samples considered in the manual annotation process have been randomly selected, while aiming for a balanced distribution, for both the dialect identification and the categorization by topic sub-tasks. Thus, a total of 120 samples have been selected, from which 60 were written in Romanian and the other 60 originated in news reports from the Republic of Moldova. For each dialect, we considered 10 samples from each of the 6 categories available in MOROCO: \emph{culture}, \emph{finance}, \emph{politics}, \emph{science}, \emph{sports} and \emph{tech}. 

Another fact about the data set is that the samples considered for annotation contain only the first sentence of the original news articles. This made the task more challenging from a human perspective, as we took away most of the context from the examples, with useful linguistic and semantic clues that could have provided a great help in inferring the correct classes. However, in the same time, our aim was to reduce the annotation time by as much as possible, since all human subjects were volunteers providing the annotations for free. As we seek to fairly compare the human skills with the performance of the ML models in differentiating among dialects, we consider the results reported in the third evaluation scenario, in which the models are trained on sentences and tested on tweets. In order for the evaluation to happen in similar circumstances, the named entities in the samples presented to the human annotators have been replaced with the special token \emph{\$ne\$}, just as in the data samples used to train and evaluate the ML models.

The summary of the human annotation is presented in Table~\ref{tab_results_human}. 
For dialect identification, the worst results are just below random chance, the accuracy of annotators \#A1 and \#A4 being $48.3\%$. Moreover, the accuracy averaged over all annotators ($53.2\%$) merely exceeds the probability of a coin toss. With an accuracy of $58.3\%$, annotator \#A3 is the only one getting closer to the results reported for the ML models in Table~\ref{tab_results_dialect}. We believe it is fair to compare the human performance at the sentence level with the performance of ML models applied on tweets. We hereby note that the accuracy of the best human annotator exceeds the accuracy of LSTM and HAN. However, SVM and KRR provide accuracy rates and macro $F_1$ scores that are about $10\%$ higher than those of annotator \#A3. The ensemble based on stacking is even better. This high difference between the ML models and the Romanian and Moldavian speaking annotators indicates that there are some subtle patterns undetected by humans. In order to discover these patterns, in Section~\ref{sec_Discussion}, we analyze Grad-CAM visualizations pointing out what models, particularly the character-level CNN, focus on.

Evaluating the human annotations for the categorization by topic task, we observe that the annotators are much better at discriminating between the six topics than identifying the dialect, the accuracy rates being between $63.3\%$ and $75.0\%$ and the macro $F_1$ scores being between $62.7\%$ and $74.8\%$. These results are comparable to the ones obtained by the ML models in the intra-dialect and cross-dialect categorization experiments presented in Tables~\ref{tab_results_categorization_intra} and~\ref{tab_results_categorization_cross}, respectively. The previous statement is specifically valid for the results reported in the second scenario, in which models are trained and tested at the sentence level.

\begin{figure}
\centering
\begin{tabular}{@{}c@{}} 
  \includegraphics[width=0.495\linewidth]{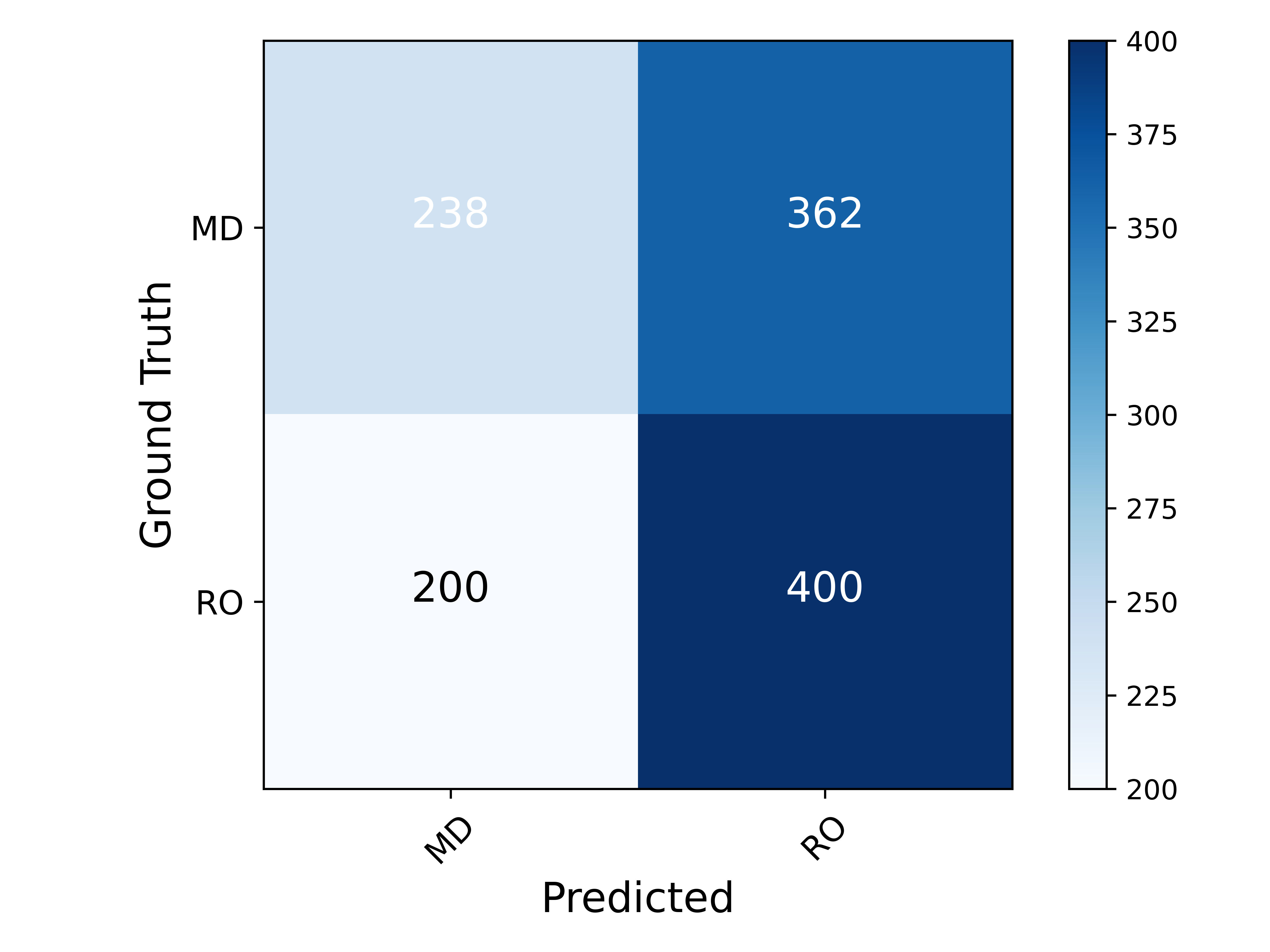} \\
    \small (a)
\end{tabular}
\begin{tabular}{@{}c@{}} 
  \includegraphics[width=0.495\linewidth]{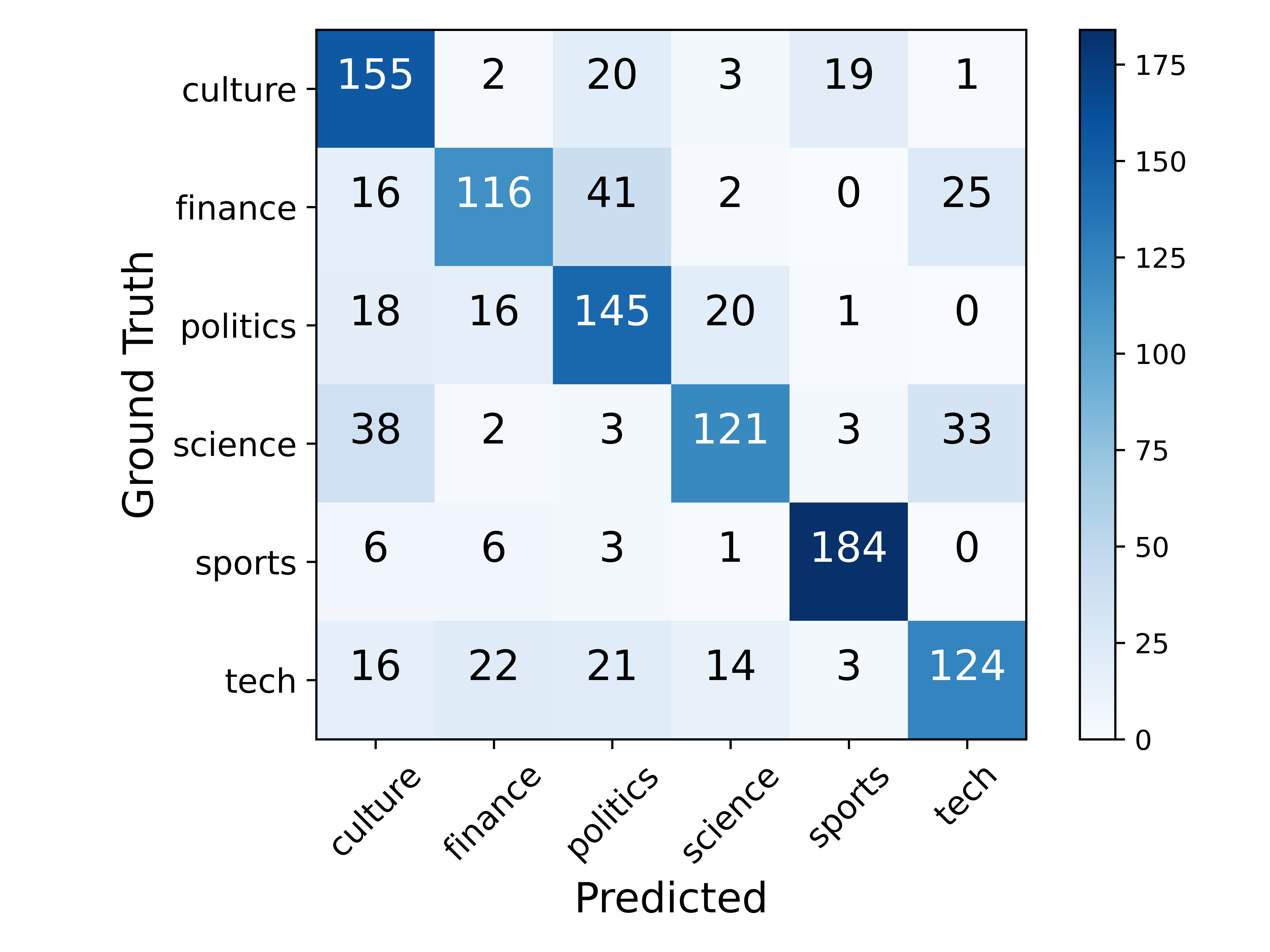} \\
    \small (b)
  \label{fig:Ng2}
\end{tabular}
\caption[Confusion matrices summed]{(a) Sum of dialect identification confusion matrices computed on all annotators included in the human evaluation study. (b) Sum of categorization by topic confusion matrices computed on all annotators included in the human evaluation study. Best viewed in color.}
\label{fig_confusion_humans}
\end{figure}

\begin{table}[t!]
\caption{Examples of sentences with ground-truth labels, as well as labels assigned by humans, for the dialect identification task. Corresponding English translations are also provided for a better comprehension.}
\label{tab_dialect_annotation_examples}
\centering
\begin{tabular}{cccccccccccc}
\hline
{{Sample ID}}   & {{Ground-Truth}} & \multicolumn{10}{c}{Labels by Annotators}\\
\cline{3-12}
                & {{Label}} & {{\#A1}} & {{\#A2}} & {{\#A3}} & {{\#A4}} & {{\#A5}} & {{\#A6}} & {{\#A7}} & {{\#A8}} & {{\#A9}} & {{\#A10}} \\
\hline 
\#S1 &  \textbf{MD} & RO & RO & RO & RO & RO & RO & RO & RO & RO & RO \\
\#S2 &  \textbf{MD} & MD & MD & MD & MD & MD & MD & MD & MD & MD & MD \\
\#S3 & \textbf{MD} & MD & MD & RO & MD & RO & RO & MD & MD & RO & RO \\
\#S4 & \textbf{RO} & RO & RO & RO & RO & RO & MD & RO & RO & RO & RO \\
\#S5 & \textbf{RO} & MD & RO & MD & MD & RO & RO & RO & MD & RO & MD \\
\#S6 & \textbf{RO} & MD & MD & MD & MD & RO & MD & RO & MD & MD & MD \\
\hline
\end{tabular}

\centering
\begin{tabular}{m{0.12\columnwidth}m{0.4\columnwidth}m{0.38\columnwidth}}
\rowcolor{gray!0!white}
\multicolumn{3}{c}{} \\
\hline
{{Sample ID}} &  {Sample} & {{English Translation}} \\
\hline 
\rowcolor{gray!20!white}
\hspace{0.6cm}\#S1 & \textit{``oamenii de \c{s}tiin\c{t}\u{a} de la \$ne\$ din \$ne\$ elaboreaz\u{a} pantaloni inteligen\c{t}i cu mu\c{s}chi artificiali, care vor oferi un sprijin suplimentar persoanelor cu mobilitatea piciorului afectat\u{a},  noteaz\u{a} \$ne\$ \$ne\$ citat de \$ne\$.''} & \textit{``scientists from \$ne\$ in \$ne\$ are fabricating smart pants with artificial muscle, which are going to help people with reduced leg mobility, says \$ne\$ \$ne\$ cited by \$ne\$.''} \\
\hspace{0.6cm}\#S2 & \textit{``leul moldovenesc se depreciaz\u{a} fa\c{t}\u{a} de moneda unic\u{a} european\u{a}.''} & \textit{``the Moldavian Leu depreciates compared to the unique European currency.''} \\
\rowcolor{gray!20!white}
\hspace{0.6cm}\#S3 & \textit{``conversa\c{t}iile incomode cu prieteni \c{s}i contacte de afaceri de alt\u{a} na\c{t}ionalitate vor fi de domeniul trecutului pentru utilizatorii serviciului \$ne\$ \$ne\$ aplica\c{t}ia pentru mobil urm\^{a}nd s\u{a} propun\u{a} traducerea \^{i}n timp real a schimburilor de mesaje.''} & \textit{``awkward conversations with friends and business contacts of another nationality will become obsolete for the \$ne\$ \$ne\$ service users as their mobile application will soon propose a real-time translation for message exchanges.'' } \\
\hspace{0.6cm}\#S4 & \textit{``acestea sunt fondurile europene pe care \$ne\$ le pierde definitiv doar cu programul \$ne\$.''} & \textit{``these are the European funds lost forever by \$ne\$ only with the \$ne\$ program.''} \\
\rowcolor{gray!20!white}
\hspace{0.6cm}\#S5 & \textit{``o sal\u{a} a teatrului \$ne\$ \$ne\$ din \$ne\$ s-a f\u{a}cut scrum.''} & \textit{``a hall of the  \$ne\$ \$ne\$ theater from \$ne\$ turned to ashes.''} \\
\hspace{0.6cm}\#S6 & \textit{``un bilet amoros r\u{a}t\u{a}cit, amenin\c{t}\u{a}ri cu sinuciderea, gesturi extreme, dictate de pasiuni la fel de extreme, o sticlu\c{t}\u{a} cu <<vitrion englezesc>>, toate acestea \^{i}nv\u{a}lm\u{a}\c{s}indu-se pe scen\u{a}.''}  & \textit{``a lost love note, suicide threats, extreme gestures, dictated by equally extreme passions, a bottle of <<English vitrion>>, all these mixing up on the stage.''} \\
\hline
\end{tabular}
\end{table}

Figure~\ref{fig_confusion_humans} (a) shows the sum of confusion matrices computed on all the annotators included in our human evaluation study, for the dialect identification task. We note that the annotators were predisposed at labeling the received samples as belonging to the Romanian dialect. On average, $63.5\%$ of the 120 samples received by each annotator have been labeled as being written in Romanian. From these, almost half are mislabeled, actually belonging to the Moldavian dialect. This predisposition can be explained by the fact that nine out of ten annotators were native Romanian speakers, hence the bias towards labeling more samples as Romanian, unless they found clues indicating otherwise. Additionally, the poor results confirm the difficulty of this binary classification task, from a human perspective.

Figure~\ref{fig_confusion_humans} (b) displays the sum of confusion matrices computed on the ten human annotators, for the categorization by topic task. For the \emph{sports} category, annotators were able to correctly classify almost all sentences, with an average of 1.6 false negatives per annotator. Since \emph{sports} is less related to the other categories and sports news likely contain semantic clues right from the first sentence regarding the category of the content, it seems natural for people to find it more distinctive.
Not the same stands for categories such as \emph{finance}, \emph{politics}, \emph{science} or \emph{tech}. Indeed, the highest confusions are between \emph{finance} and \emph{politics} and between \emph{science} and \emph{tech}, respectively.

In Table~\ref{tab_dialect_annotation_examples}, we display six samples selected from the data set provided to the human annotators. Among the presented samples, the first three belong to the Moldavian dialect, while the last three belong to the Romanian dialect. For a better comprehension, the English translation of each sample is also included in Table~\ref{tab_dialect_annotation_examples}. We selected the samples considering three different cases: $(d.i)$ most annotators agree on the label, but the plurality vote label does not match the ground-truth label; $(d.ii)$ most annotators agree on a label that matches the ground-truth label; $(d.iii)$ there are strong disagreements among annotators, such that a majority cannot be determined. The first and the sixth rows in Table~\ref{tab_dialect_annotation_examples} are representative for case $(d.i)$. In sample \#S1, there is no linguistic or semantic clue to indicate that the sentence belongs to the Romanian dialect, yet all annotators made this choice, against the ground-truth label (Moldavian). One explanation for this choice is perhaps motivated by the fact that Romania is a more developed country from a scientific point of view. Hence, the annotators might have been biased in their belief that a news article talking about scientists is more likely to come from Romania than from the Republic of Moldova. Sample \#S6 contains several words that are not commonly used in the Romanian language, hence, all annotators decide to label it as belonging to the Moldavian dialect. Samples \#S2 and \#S4 are representative for case $(d.ii)$, most of the votes matching the correct label. Sample \#S2 contains an explicit clue suggesting that it belongs to the Moldavian dialect, namely the adjective ``moldovenesc'' when referring to the currency used in the Republic of Moldova. Similarly, in sentence \#S4, the clue is the noun phrase ``fondurile europene''. The Republic of Moldova is not a member of the European Union. Thus, it becomes clear for anyone who knows this information that the corresponding sentence is more likely to originate in Romania, as Romania is involved in receiving funds from the European Union. Finally, samples \#S3 and \#S5 are representative for case $(d.iii)$. We notice that, in sample \#S5, there is simply not enough context to infer the dialect, while sample \#S3 does not bare any clues to indicate the dialect, although the sentence is longer. Interestingly, in the presented samples neither we nor the annotators were able to spot any dialectal clues. Although some samples were labeled correctly, the clues indicating the correct dialect are more related to the subject rather than the dialect. Until this point, we conclude that either the dialectal patterns are missing or they are very hard to spot by humans. The analysis provided in Section~\ref{sec_Discussion} reveals that the character-level CNN does learn some interesting dialectal clues, which we were not aware of.

\begin{table}[tp]
\caption{Examples of sentences with ground-truth labels, as well as labels assigned by humans, for the categorization by topic task. Corresponding English translations are also provided for a better comprehension. The following abbreviations are used for the six topics: CUL -- \emph{culture}, FIN -- \emph{finance}, POL -- \emph{politics}, SCI -- \emph{science}, SPO -- \emph{sports} and TEC -- \emph{tech}.}
\label{tab_category_annotation_examples}
\centering
\footnotesize
\begin{tabular}{cccccccccccc}
\hline
{{Sample ID}}   & {{Ground-Truth}} & \multicolumn{10}{c}{Labels by Annotators}\\
\cline{3-12}
                & {{Label}} & {{\#A1}} & {{\#A2}} & {{\#A3}} & {{\#A4}} & {{\#A5}} & {{\#A6}} & {{\#A7}} & {{\#A8}} & {{\#A9}} & {{\#A10}}\\
\hline
\#S7 & \textbf{CUL} & CUL & CUL & CUL & CUL & CUL & CUL & CUL & CUL & CUL & CUL \\
\#S8 & \textbf{CUL} & POL & SPO & CUL & POL & SPO & SPO & CUL & POL & POL & SPO \\
\#S9 & \textbf{FIN} & FIN & FIN & FIN & FIN & FIN & FIN & FIN & FIN & FIN & FIN \\
\#S10  & \textbf{FIN} & POL & POL & POL & FIN & POL & POL & POL & POL & FIN & POL \\
\#S11 & \textbf{POL} & POL & POL & POL & POL & POL & POL & POL & POL & POL & POL  \\
\#S12 & \textbf{POL} & POL & POL & POL & POL & POL & CUL & POL & POL & POL & POL \\
\#S13 & \textbf{SCI} & SCI & SCI & SCI & SCI & SCI & SCI & SCI & SCI & SCI & SCI \\
\#S14  & \textbf{SCI} & CUL & CUL & SCI & TEC & TEC & TEC & SCI & SCI & SCI & SCI \\
\#S15 & \textbf{SPO} & SPO & SPO & SPO & SPO & SPO & SPO & SPO & SPO & SPO & SPO \\
\#S16  & \textbf{SPO} & SPO & SPO & CUL & SPO & SPO & SPO & CUL & SPO & SPO & CUL \\
\#S17 & \textbf{TEC} & TEC & TEC & TEC & TEC & TEC & TEC & TEC & TEC & TEC & TEC \\
\#S18 & \textbf{TEC} & TEC & TEC & FIN & FIN & TEC & FIN & TEC & TEC & FIN & FIN \\
\hline
\end{tabular}

\centering
\footnotesize
\begin{tabular}{m{0.11\columnwidth}m{0.4\columnwidth}m{0.39\columnwidth}}
\rowcolor{gray!0!white}
\multicolumn{3}{c}{}
\vspace{-0.3cm}\\
\hline
\hspace{0.1cm}{{Sample ID}} &  \textbf{Sample} & {\textbf{English Translation}} \\
\hline 
\rowcolor{gray!20!white}
\hspace{0.6cm}\#S7 & \textit{``\$ne\$ \$ne\$ recunoscut pentru felul lui unic de a fi, revine \^{i}n \$ne\$ cu tolba plin\u{a} de muzic\u{a} \c{s}i poezie.''} & \textit{``\$ne\$ \$ne\$ famous for his unique way of being, comes back in \$ne\$ with lots of music and poesy.''}\\
\hspace{0.6cm}\#S8 & \textit{``multe voci sustin ca patruderea puternica a \$ne\$ si a \$ne\$ pe podiumul \$ne\$ intareste si mai mult ideea ca aceasta competitie nu este decat una (geo) politica.''} & \textit{``several voices argue that \$ne\$ and \$ne\$'s strong infiltration on the \$ne\$ podium further confirms the idea that this is nothing more than a (geo) political competition.''}\\
\rowcolor{gray!20!white}
\hspace{0.6cm}\#S9 & \textit{``locurile de munc\u{a} bine pl\u{a}tite pot fi g\u{a}site \c{s}i \^{i}n alte sectoare ale economiei, mai pu\c{t}in populare, \^{i}n care b\u{a}t\u{a}lia este mai mic\u{a}.''} & \textit{``well paid work places can also be found in other less popular sectors of the economy, with less competition.''}\\
\hspace{0.6cm}\#S10 & \textit{``coali\c{t}ia la guvernare spune c\u{a} atunci c\^{a}nd accizele au fost reduse, la 1 ianuarie, carburan\c{t}ii nu s-au ieftinit.''} & \textit{``governing coalition states that fuel has not become cheaper with the reduction of excise duties that happened on January 1st.''} \\
\rowcolor{gray!20!white}
\hspace{0.6cm}\#S11 & \textit{``dac\u{a} duminica viitoare ar avea loc alegeri parlamentare, \^{i}n \$ne\$ ar accede trei partide: \$ne\$ \c{s}i \$ne\$.''} & \textit{``if there would be parliamentary elections next Sunday, in \$ne\$ three parties would accede: \$ne\$ and \$ne\$.''} \\
\hspace{0.6cm}\#S12 & \textit{``\$ne\$ \$ne\$ \c{s}i \$ne\$ ( \$ne\$ ) \c{s}i \$ne\$ \$ne\$ \c{s}i \$ne\$ (\$ne\$) au devenit ast\u{a}zi membri observatori ai \$ne\$ \$ne\$ \$ne\$ ( \$ne\$ ).''} & \textit{``\$ne\$ \$ne\$ and \$ne\$ ( \$ne\$ ) and \$ne\$ \$ne\$ and \$ne\$ (\$ne\$) are, as of today, observer members of \$ne\$ \$ne\$ \$ne\$ ( \$ne\$ ).''} \\
\rowcolor{gray!20!white}
\hspace{0.6cm}\#S13 & \textit{``\$ne\$ a anun\c{t}at descoperirea unui sistem solar asem\u{a}n\u{a}tor cu al nostru, care are opt planete.''} & \textit{\$ne\$ has announced the discovery of a new solar system similar to ours, which has eight planets.''} \\
\hspace{0.6cm}\#S14 & \textit{``totul se intampla la bordul \$ne\$ \$ne\$ \$ne\$.''} & \textit{``everything happens on board of \$ne\$ \$ne\$ \$ne\$.''} \\
\rowcolor{gray!20!white}
\hspace{0.6cm}\#S15 & \textit{``ronaldinho s-a retras oficial, anun\c{t}ul venind din partea fratelui acestuia, cel care-i este \c{s}i agent.''} & \textit{``Ronaldinho has officially retired, announces his brother, who is also his agent.''} \\
\hspace{0.6cm}\#S16 & \textit{``dupa ce s-a dat cu motorul in favelas din \$ne\$ pe un roller coaster din \$ne\$ \$ne\$ dar si la \$ne\$ \$ne\$ motociclistul francez care calatoreste in toata lumea cautand cele mai spectaculoase locuri pentru trial freestyle, \$ne\$ \$ne\$ a revenit in \$ne\$ pentru un proiect inedit.''} & \textit{``after he has been biking in favelas from \$ne\$ on a roller coaster from \$ne\$ \$ne\$ and also in \$ne\$ \$ne\$, the French biker who travels the world seeking the most spectacular places for freestyle trial, \$ne\$ \$ne\$ came back in \$ne\$ for a novel project.''} \\
\rowcolor{gray!20!white}
\hspace{0.6cm}\#S17 & \textit{``telefoanele inteligente au ajuns s\u{a} \^{i}i traduc\u{a} pe bebelu\c{s}i.''} & \textit{``smartphones got to the point where they can translate babies' language.''}\\
\hspace{0.6cm}\#S18 & \textit{``pe pia\c{t}a online a \$ne\$.''} & \textit{``on the online market of \$ne\$.''}\\
\hline
\end{tabular}
\end{table}

In Table~\ref{tab_category_annotation_examples}, we present sentences with category labels for two different cases: $(c.i)$ the correct category is chosen in unanimity; $(c.ii)$ there are disagreements among annotators, regardless of the final result of plurality voting. Each of these two cases is exemplified through one sentence for each of the six categories. Samples \#S7, \#S9, \#S11, \#S13, \#S15, \#S17 are representative for case $(c.i)$. The nouns ``muzic\u{a}'' and ``poezie'' in example \#S7 are strong clues for the \emph{culture} category, hence the unanimity of votes in this direction. In sample \#S9, the keyword ``economie'' gives the strongest clue for the \emph{finance} category, while in sample \#S11, the noun phrase ``alegeri parlamentare'' suggests that the sentence belongs to the \emph{politics} category. However, sample \#S13 does not seem to contain any specific phrase that can be considered a strong indicator for the \emph{science} topic. Here, it is the entire context that reveals the nature of the sentence. The name of a famous football player has escaped our named entity removal process, representing the reason why sentence \#S15 was unanimously classified as belonging to the \emph{sports} topic. In example \#S17, the construct ``telefoane inteligente'', which translates to ``smartphones'', is a very strong indicator for the \emph{tech} category. Therefore, the annotators unanimously labeled \#S17 as part of the \emph{tech} sector. Examples \#S8, \#S10, \#S12, \#S14, \#S16, \#S18 are representative for case $(c.ii)$, having at least one wrong label among the manual annotations provided by the ten annotators. Only two out of ten annotators have correctly labeled sample \#S8 as belonging to the \emph{culture} topic. The other annotators were deceived by the fact that sample \#S8 contains the word ``politica'', suggesting the \emph{politics} label, or the word ``podium'', suggesting the \emph{sports} label. The annotations of sample \#S10 confirm the confusion between \emph{finance} and \emph{politics} observed in the confusion matrix depicted in Figure~\ref{fig_confusion_humans}~(b). In sample \#S10, we observe a reason suggesting that the label is \emph{politics}, namely the presence of the noun phrase ``coali\c{t}ia la guvernare''. If the annotators would have considered the noun ``accizele'' (taxes) as more relevant, they would have been able to find the correct category, i.e.~\emph{finance}. Sample \#S12 contains very few words along with many placeholders for named entities. However, most of the annotators know that ``membri observatori'' is a political function inside the European Union, hence the label \emph{politics}. Sample \#S14 presents strong disagreements among the annotators. This is expected due to the very short sentence lacking sufficient context to label the example. Misclassified \emph{sports} samples were very few in the data set, as we can also see in Figure~\ref{fig_confusion_humans}~(b). Sample \#S16 is one of the few where three annotators did not mark the text as belonging to the \emph{sports} category. Leaving aside the lack of context in sample \#S18, we note that the noun phrase ``pia\c{t}a online'' (online market) might suggest the \emph{finance} and the \emph{tech} topics. The labels provided by the annotators are divided between these two topics, confirming our hypothesis about ``pia\c{t}a online''.

\section{Discussion}
\label{sec_Discussion}

So far, it remains unclear if there are any dialectal clues in the news articles from MOROCO. One hypothesis (H1) is that there are no dialectal clues, since Romanian speakers had a hard time distinguishing between the two dialects, as shown in Table~\ref{tab_results_human}. In this case, the good performance of the machine learning models can be explained through other factors, e.g.~subjects specific to each of the two countries. The alternative hypothesis (H2) is that the samples contain dialectal clues, since the machine learning models trained on news articles are able to classify tweets collected from a different time period. In this case, the low performance of human annotators can be explained if we consider that the dialectal clues are harder to spot than expected. In order to find out which hypothesis is valid, we analyze the discriminative features learned by the character-level CNN, which is among the top three individual dialect identification systems. We opted for the character-level CNN in favor of the better SVM and KRR, as it allows us to look at discriminative features using Grad-CAM, a technique that was initially used to explain decisions of convolutional neural networks applied on images~\cite{Selvaraju-ICCV-2017}. We adapted this technique for the character-level CNN, presenting the corresponding visualizations in Tables~\ref{tab_GradCAM_RO} and~\ref{tab_GradCAM_MD}, respectively.

\begin{table}[!th]
\caption{Grad-CAM visualizations for the character-level CNN applied on Romanian samples. The shade of blue indicates the importance of the group of characters, i.e.~darker shades highlight more important features and lighter shades indicate less important features. For a better reading, spaces are not highlighted. Best viewed in color.}
\label{tab_GradCAM_RO}
\footnotesize
\begin{tabular}{m{0.04\columnwidth} m{0.48\columnwidth} m{0.38\columnwidth}}
\hline
{\textbf{ID}} &  \textbf{Visualization} & \textbf{English Translation} \\
\hline 

\#R1 & \textcolor{white}{\hlc[blue1]{ford}}\textcolor{white}{\hlc[blue1]{a}}\textcolor{white}{\hlc[blue1]{demarat}}\textcolor{white}{\hlc[blue1]{produc\c{t}ia}}\textcolor{black}{\hlc[blue6]{noului}}\textcolor{black}{\hlc[blue6]{s\u{a}u}}\textcolor{black}{\hlc[blue6]{model}}\textcolor{black}{\hlc[blue8]{\$ne\$}}\textcolor{black}{\hlc[blue8]{la}}\textcolor{black}{\hlc[blue8]{uzina}}\textcolor{black}{\hlc[blue8]{din}}\textcolor{white}{\hlc[blue4]{\$ne\$}}\textcolor{white}{\hlc[blue4]{.}} & \textit{``Ford has started the production of the new model \$ne\$ at their factory in \$ne\$.''}\\

\#R2 & \textcolor{white}{\hlc[blue1]{mai}}\textcolor{white}{\hlc[blue1]{pe}}\textcolor{white}{\hlc[blue1]{rom\^ane\c{s}te,}}\textcolor{white}{\hlc[blue1]{taie}}\textcolor{white}{\hlc[blue1]{frunze}}\textcolor{white}{\hlc[blue1]{la}}\textcolor{white}{\hlc[blue1]{c\^aini}}\textcolor{black}{\hlc[blue6]{.}} & \textit{``as the romanian saying goes, he's cutting leaves to the dogs.''} \\

\#R3 & \textcolor{white}{\hlc[blue1]{a\c{s}a}}\textcolor{white}{\hlc[blue1]{au}}\textcolor{white}{\hlc[blue1]{ap\u{a}rut,}}\textcolor{white}{\hlc[blue1]{\^in}}\textcolor{white}{\hlc[blue1]{g\u{a}ri,}}\textcolor{black}{\hlc[blue6]{tonomatele}}\textcolor{black}{\hlc[blue6]{cu}}\textcolor{black}{\hlc[blue6]{tot}}\textcolor{white}{\hlc[blue3]{felul}}\textcolor{white}{\hlc[blue3]{de}}\textcolor{white}{\hlc[blue3]{publica\c{t}ii,}}\textcolor{white}{\hlc[blue1]{contra}}\textcolor{white}{\hlc[blue1]{cost,}}\textcolor{white}{\hlc[blue1]{sau}}\textcolor{white}{\hlc[blue1]{bibliotecile}}\textcolor{white}{\hlc[blue1]{pentru}}\textcolor{white}{\hlc[blue4]{corporati\c{s}ti,}}\textcolor{white}{\hlc[blue4]{cu}}\textcolor{white}{\hlc[blue3]{livrare}}\textcolor{white}{\hlc[blue3]{direct}}\textcolor{white}{\hlc[blue3]{la}}\textcolor{white}{\hlc[blue3]{birou}}\textcolor{white}{\hlc[blue3]{.}} & \textit{``this is how there have appeared, in train stations, jukeboxes with all kinds of books, requiring payment, or libraries for corporate people, with delivery directly at the office.''} \\

\#R4 & \textcolor{white}{\hlc[blue5]{\c{s}eful}}\textcolor{white}{\hlc[blue5]{\$ne\$}}\textcolor{white}{\hlc[blue5]{\$ne\$}}\textcolor{white}{\hlc[blue5]{\$ne\$}}\textcolor{black}{\hlc[blue7]{\c{s}i}}\textcolor{black}{\hlc[blue7]{adjunctul}}\textcolor{black}{\hlc[blue7]{acestuia,}}\textcolor{black}{\hlc[blue7]{\$ne\$}}\textcolor{black}{\hlc[blue7]{\$ne\$}}\textcolor{black}{\hlc[blue7]{s}}\textcolor{black}{\hlc[blue7]{-}}\textcolor{white}{\hlc[blue5]{au}}\textcolor{white}{\hlc[blue5]{prezentat,}}\textcolor{white}{\hlc[blue5]{joi,}}\textcolor{black}{\hlc[blue6]{la}}\textcolor{black}{\hlc[blue6]{\$ne\$}}\textcolor{black}{\hlc[blue6]{\$ne\$}}\textcolor{black}{\hlc[blue6]{\^in}}\textcolor{black}{\hlc[blue6]{dosarul}}\textcolor{white}{\hlc[blue5]{violen\c{t}elor}}\textcolor{white}{\hlc[blue5]{de}}\textcolor{white}{\hlc[blue1]{la}}\textcolor{white}{\hlc[blue1]{mitingul}}\textcolor{white}{\hlc[blue1]{din}}\textcolor{white}{\hlc[blue1]{10}}\textcolor{white}{\hlc[blue2]{august}}\textcolor{white}{\hlc[blue2]{din}}\textcolor{white}{\hlc[blue2]{\$ne\$}}\textcolor{white}{\hlc[blue2]{\$ne\$}}\textcolor{white}{\hlc[blue5]{.}} & \textit{``the head of \$ne\$ \$ne\$ \$ne\$ and his assistant were present, on Thursday, at \$ne\$ \$ne\$ in the criminal case of the violence acts that happened at the protests on August 10th from \$ne\$ \$ne\$.''} \\

\#R5 & \textcolor{white}{\hlc[blue2]{compania}}\textcolor{white}{\hlc[blue2]{\^in}}\textcolor{white}{\hlc[blue2]{cauz\u{a}}}\textcolor{white}{\hlc[blue2]{nu}}\textcolor{white}{\hlc[blue5]{mai}}\textcolor{white}{\hlc[blue5]{vindea}}\textcolor{white}{\hlc[blue5]{modele}}\textcolor{white}{\hlc[blue5]{diesel}}\textcolor{white}{\hlc[blue5]{pe}}\textcolor{white}{\hlc[blue5]{pia\c{t}a}}\textcolor{white}{\hlc[blue5]{american\u{a}}}\textcolor{black}{\hlc[blue6]{din}}\textcolor{black}{\hlc[blue6]{2015}}\textcolor{black}{\hlc[blue6]{.}}
 & \textit{``the said company hasn't sold diesel models on the American market since 2015.''} \\
 
 \#R6 & \textcolor{black}{\hlc[blue6]{partidele}}\textcolor{black}{\hlc[blue6]{au}}\textcolor{black}{\hlc[blue6]{o}}\textcolor{black}{\hlc[blue6]{nou\u{a}}}\textcolor{black}{\hlc[blue6]{tem\u{a}}}\textcolor{black}{\hlc[blue6]{de}}\textcolor{black}{\hlc[blue6]{campanie}}\textcolor{black}{\hlc[blue6]{-}}\textcolor{white}{\hlc[blue1]{definirea}}\textcolor{white}{\hlc[blue1]{familiei}}\textcolor{white}{\hlc[blue3]{.}}  & \textit{``the parties have a new campaign theme - defining family.''} \\

\#R7 & \textcolor{white}{\hlc[blue1]{coali\c{t}ia}}\textcolor{white}{\hlc[blue1]{la}}\textcolor{white}{\hlc[blue1]{guvernare}}\textcolor{white}{\hlc[blue3]{spune}}\textcolor{white}{\hlc[blue3]{c\u{a}}}\textcolor{white}{\hlc[blue3]{atunci}}\textcolor{white}{\hlc[blue3]{c\^and}}\textcolor{black}{\hlc[blue7]{accizele}}\textcolor{black}{\hlc[blue7]{au}}\textcolor{black}{\hlc[blue7]{fost}}\textcolor{black}{\hlc[blue10]{reduse,}}\textcolor{black}{\hlc[blue10]{la}}\textcolor{black}{\hlc[blue10]{1}}\textcolor{black}{\hlc[blue10]{ianuarie,}}\textcolor{black}{\hlc[blue9]{carburan\c{t}ii}}\textcolor{black}{\hlc[blue9]{nu}}\textcolor{white}{\hlc[blue5]{s}}\textcolor{white}{\hlc[blue5]{-}}\textcolor{white}{\hlc[blue5]{au}}\textcolor{white}{\hlc[blue5]{ieftinit}}\textcolor{white}{\hlc[blue5]{.}} & \textit{``governing coalition states that  fuel  has  not become  cheaper with the reduction of excise duties that happened on January 1st.''}\\

\#R8 & \textcolor{white}{\hlc[blue3]{bancherii}}\textcolor{white}{\hlc[blue3]{nu}}\textcolor{white}{\hlc[blue3]{stau}}\textcolor{white}{\hlc[blue3]{cu}}\textcolor{white}{\hlc[blue1]{mainile}}\textcolor{white}{\hlc[blue1]{in}}\textcolor{white}{\hlc[blue1]{san}}\textcolor{white}{\hlc[blue1]{.}} &  \textit{``bankers do not sit with the hands on their chests.''} \\

\#R9 & \textcolor{black}{\hlc[blue6]{hidrologii}}\textcolor{black}{\hlc[blue6]{au}}\textcolor{black}{\hlc[blue6]{emis,}}\textcolor{white}{\hlc[blue5]{s\^amb\u{a}t\u{a},}}\textcolor{white}{\hlc[blue5]{mai}}\textcolor{white}{\hlc[blue5]{multe}}\textcolor{white}{\hlc[blue5]{avertiz\u{a}ri}}\textcolor{white}{\hlc[blue5]{cod}}\textcolor{white}{\hlc[blue5]{galben}}\textcolor{white}{\hlc[blue2]{de}}\textcolor{white}{\hlc[blue2]{inunda\c{t}ii,}}\textcolor{white}{\hlc[blue2]{scurgeri}}\textcolor{white}{\hlc[blue3]{de}}\textcolor{white}{\hlc[blue3]{pe}}\textcolor{white}{\hlc[blue3]{versan\c{t}i,}}\textcolor{white}{\hlc[blue3]{toren\c{t}i}}\textcolor{white}{\hlc[blue3]{\c{s}i}}\textcolor{white}{\hlc[blue3]{p\^araie,}}\textcolor{white}{\hlc[blue1]{valabile}}\textcolor{white}{\hlc[blue1]{pentru}}\textcolor{black}{\hlc[blue6]{r\^auri}}\textcolor{black}{\hlc[blue6]{din}}\textcolor{black}{\hlc[blue6]{\c{s}ase}}\textcolor{black}{\hlc[blue6]{jude\c{t}e}}\textcolor{black}{\hlc[blue9]{.}} & \textit{``hydrologists have emitted, on Saturday, more yellow code flood warnings, 
runoff from slopes, torrents and streams, valid for rivers in six counties.''} \\

\#R10 & \textcolor{white}{\hlc[blue1]{\$ne\$}}\textcolor{white}{\hlc[blue1]{a}}\textcolor{white}{\hlc[blue1]{precizat}}\textcolor{white}{\hlc[blue1]{c\u{a}}}\textcolor{white}{\hlc[blue1]{este}}\textcolor{white}{\hlc[blue4]{vorba}}\textcolor{white}{\hlc[blue4]{despre}}\textcolor{white}{\hlc[blue4]{semnarea}}\textcolor{white}{\hlc[blue5]{unui}}\textcolor{white}{\hlc[blue5]{acord}}\textcolor{white}{\hlc[blue5]{\^intre}}\textcolor{black}{\hlc[blue6]{o}}\textcolor{black}{\hlc[blue6]{firm\u{a}}}\textcolor{black}{\hlc[blue6]{privat\u{a}}}\textcolor{white}{\hlc[blue4]{rom\^aneasc\u{a}}}\textcolor{white}{\hlc[blue4]{\c{s}i}}\textcolor{white}{\hlc[blue4]{una}}\textcolor{white}{\hlc[blue3]{dintre}}\textcolor{white}{\hlc[blue3]{cele}}\textcolor{white}{\hlc[blue3]{mai}}\textcolor{white}{\hlc[blue3]{mari}}\textcolor{white}{\hlc[blue1]{companii}}\textcolor{white}{\hlc[blue1]{din}}\textcolor{white}{\hlc[blue1]{lume}}\textcolor{black}{\hlc[blue6]{-}}\textcolor{black}{\hlc[blue6]{o}}\textcolor{black}{\hlc[blue6]{firm\u{a}}}\textcolor{black}{\hlc[blue6]{american\u{a}}}\textcolor{black}{\hlc[blue8]{de}}\textcolor{black}{\hlc[blue8]{armament,}}\textcolor{black}{\hlc[blue8]{care}}\textcolor{white}{\hlc[blue4]{produce,}}\textcolor{white}{\hlc[blue4]{printre}}\textcolor{white}{\hlc[blue4]{altele,}}\textcolor{white}{\hlc[blue4]{\c{s}i}}\textcolor{white}{\hlc[blue4]{celebrele}}\textcolor{black}{\hlc[blue8]{rachete}}\textcolor{black}{\hlc[blue8]{\$ne\$}}\textcolor{black}{\hlc[blue8]{.}} & \textit{``\$ne\$ has stated that this is about signing an agreement between a private Romanian company and one of the biggest companies world wide - an American weapons business, which manufactures, among others, the famous rockets \$ne\$.''}\\
\hline
\end{tabular}
\end{table}

We quantized the importance of each character using $10$ shades of blue (for Romanian) or $10$ shades of red (for Moldavian), the darker shades representing more relevant features and the lighter shades representing less relevant features, respectively. In order to extract the importance of each character, we used the weights learned by the last convolutional layer in the network as well as the spatial localization kept in the activation maps resulted upon convolving filters of predefined size over the input fed to the model. In the remainder of this discussion, we try to explain why the features considered important by the character-level CNN also make sense from a human perspective. 

We provide a set of visualizations for Romanian sentences in Table~\ref{tab_GradCAM_RO}. In sample \#R1, the model focuses on the first four words, but the one indicating the dialect is ``demarat''. This word, which translates to ``started'', is used in Romanian to indicate that the start of a construction process. In Moldavian, the word ``\^{i}nceput'' would have probably been used instead to express the same thing. We note that the word ``\^{i}nceput'' is also commonly used in Romania, but in typically different contexts. Perhaps this is why the model also highlights the neighboring word ``produc\c{t}ia'' (production).
Sample \#R2 contains an entire Romanian proverb which is, as a whole, predictive for this dialect. It refers to people doing useless jobs, e.g.~``cutting leaves to the dogs''. In sample \#R3, the CNN focuses on two separate groups of words, but we believe the dialectal clue is the ``contra cost'' expression, which is typically used in Romanian to express the fact that some product or service is not for free, but it requires some payment from the customer.
Example \#R4 contains a type of news that has dominated the Romanian media for months, namely the protest against the Romanian government on August 10th, 2018. Therefore, the features highlighted by the CNN have no dialectal clues, except perhaps for the word ``miting'', which is preferred instead of the synonym ``protest'', the latter one being more common in the Republic of Moldova. In samples \#R5 and \#R10, the CNN focuses on the nouns ``compania'' (singular of ``company'') or ``companii'' (plural of ``company''), respectively. From our observations, in Moldavian news reports, writers use ``\^{i}ntreprindere'', while in Romanian news reports, the synonym ``companie'' is rather used. We note that ``companie'' and ``\^{i}ntreprindere'' exist in both Romanian and Moldavian, but the preference for one or the other depends on the dialect. Sample \#R6 refers to what was a really hot and controversial topic in Romania, namely that of changing the definition of family (``definirea familiei'') in the constitution of Romania. We can safely say this is not a dialectal topic. In sample \#R7, the model focuses on the noun phrase ``coali\c{t}ia la gurvernare'' (governing coalition). In the Republic of Moldova, the same concept is expressed through the noun phrase ``coali\c{t}ia de guvern\u{a}m\^{i}nt''. Sample \#R8 contains a Romanian saying, namely ``nu stau cu mainile in san'', which is used to express that the bankers took some action instead of waiting for something to happen. Sample \#R9 contains the noun ``toren\c{t}i'', which is never used in the Republic of Moldova with the meaning of \emph{weather torrent}, only with the meaning of \emph{web torrent}. The CNN also considers as relevant the word ``valabile'', which is rarely used in the Republic of Moldova. Hence, sample \#R9 contains more than one dialectal pattern. In summary, we find that the CNN discovers some interesting dialectal patterns, which we were unaware of before seeing the Grad-CAM visualizations. However, there is a small percentage of sentences, namely \#R4 and \#R6, that have no dialectal patterns, but are correctly labeled by the CNN because of the subjects that are related to events in Romania.


\begin{table}[!t]
\caption{Grad-CAM visualizations for the character-level CNN applied on Moldavian samples. The shade of blue indicates the importance of the group of characters, i.e.~darker shades highlight more important features and lighter shades indicate less important features. For a better reading, spaces are not highlighted. Best viewed in color.}
\label{tab_GradCAM_MD}
\footnotesize
\begin{tabular}{m{0.04\columnwidth} m{0.48\columnwidth} m{0.38\columnwidth}}
\hline
{\textbf{ID}} &  \textbf{Visualization} & \textbf{English Translation}  \\
\hline 

\#M1 & \textcolor{white}{\hlc[red1]{cabinetul}}\textcolor{white}{\hlc[red1]{de}}\textcolor{white}{\hlc[red1]{mini\c{s}tri}}\textcolor{white}{\hlc[red3]{a}}\textcolor{white}{\hlc[red3]{aprobat,}}\textcolor{white}{\hlc[red3]{\^in}}\textcolor{white}{\hlc[red3]{cadrul}}\textcolor{white}{\hlc[red2]{\c{s}edin\c{t}ei}}\textcolor{white}{\hlc[red2]{de}}\textcolor{white}{\hlc[red2]{ast\u{a}zi,}}\textcolor{white}{\hlc[red4]{modific\u{a}ri}}\textcolor{white}{\hlc[red4]{\c{s}i}}\textcolor{black}{\hlc[red7]{complet\u{a}ri}}\textcolor{black}{\hlc[red7]{la}}\textcolor{black}{\hlc[red7]{\$ne\$}}\textcolor{black}{\hlc[red8]{privind}}\textcolor{black}{\hlc[red8]{procedura}}\textcolor{black}{\hlc[red8]{de}}\textcolor{black}{\hlc[red9]{repatriere}}\textcolor{black}{\hlc[red9]{a}}\textcolor{black}{\hlc[red9]{copiilor}}\textcolor{black}{\hlc[red10]{\c{s}i}}\textcolor{black}{\hlc[red10]{adul\c{t}ilor}}\textcolor{black}{\hlc[red9]{victime}}\textcolor{black}{\hlc[red9]{ale}}\textcolor{black}{\hlc[red9]{traficului}}\textcolor{black}{\hlc[red9]{de}}\textcolor{black}{\hlc[red9]{fiin\c{t}e}}\textcolor{black}{\hlc[red9]{umane,}}\textcolor{black}{\hlc[red7]{traficului}}\textcolor{black}{\hlc[red7]{ilegal}}\textcolor{white}{\hlc[red2]{de}}\textcolor{white}{\hlc[red2]{migran\c{t}i,}}\textcolor{white}{\hlc[red2]{precum}}\textcolor{white}{\hlc[red3]{\c{s}i}}\textcolor{white}{\hlc[red3]{a}}\textcolor{white}{\hlc[red3]{copiilor}}\textcolor{white}{\hlc[red3]{ne\^inso\c{t}i\c{t}i}}\textcolor{white}{\hlc[red3]{.}} & \textit{``the cabinet has approved, in today's meeting, changes and completions to \$ne\$ regarding the procedure for repatriation of children and adults who are victims of human trafficking, illegal immigrants trafficking, as well as the one regarding unaccompanied children.''} \\

\#M2 & \textcolor{black}{\hlc[red7]{\$ne\$}}\textcolor{black}{\hlc[red7]{are}}\textcolor{black}{\hlc[red7]{tot}}\textcolor{black}{\hlc[red7]{ce}}\textcolor{black}{\hlc[red7]{\^ii}}\textcolor{black}{\hlc[red7]{trebuie}}\textcolor{white}{\hlc[red4]{pentru}}\textcolor{white}{\hlc[red4]{a}}\textcolor{white}{\hlc[red4]{reu\c{s}i,}}\textcolor{white}{\hlc[red4]{iar}}\textcolor{white}{\hlc[red4]{\$ne\$}}\textcolor{white}{\hlc[red4]{\$ne\$}}\textcolor{white}{\hlc[red4]{a}}\textcolor{black}{\hlc[red8]{condus}}\textcolor{black}{\hlc[red8]{\c{t}ara}}\textcolor{black}{\hlc[red8]{cu}}\textcolor{black}{\hlc[red8]{m\^ina}}\textcolor{white}{\hlc[red5]{ferm\u{a},}}\textcolor{white}{\hlc[red5]{cu}}\textcolor{white}{\hlc[red5]{minte}}\textcolor{white}{\hlc[red5]{limpede}}\textcolor{white}{\hlc[red3]{\c{s}i}}\textcolor{white}{\hlc[red3]{cu}}\textcolor{white}{\hlc[red3]{sufletul}}\textcolor{white}{\hlc[red5]{la}}\textcolor{white}{\hlc[red5]{oameni,}}\textcolor{white}{\hlc[red5]{f\u{a}c\^indu}}\textcolor{white}{\hlc[red2]{-}}\textcolor{white}{\hlc[red2]{\c{s}i}}\textcolor{white}{\hlc[red2]{datoria}}\textcolor{white}{\hlc[red2]{fa\c{t}\u{a}}}\textcolor{white}{\hlc[red1]{de}}\textcolor{white}{\hlc[red1]{oameni}}\textcolor{white}{\hlc[red1]{.}} & \textit{``\$ne\$ has everything that is needed in order to win, but \$ne\$ \$ne\$ has ruled the country with a firm hand, a clear mind and with the soul close to people, while doing his/her duty to people.''} \\

\#M3 & \textcolor{black}{\hlc[red8]{facebook}}\textcolor{black}{\hlc[red8]{\c{s}tie}}\textcolor{black}{\hlc[red8]{multe}}\textcolor{black}{\hlc[red7]{lucruri}}\textcolor{black}{\hlc[red7]{despre}}\textcolor{black}{\hlc[red7]{tine,}}\textcolor{black}{\hlc[red7]{dintre}}\textcolor{black}{\hlc[red7]{care}}\textcolor{black}{\hlc[red7]{majoritatea}}\textcolor{white}{\hlc[red3]{s\^int}}\textcolor{white}{\hlc[red3]{\^imp\u{a}rt\u{a}\c{s}ite}}\textcolor{white}{\hlc[red1]{cu}}\textcolor{white}{\hlc[red1]{prietenii}}\textcolor{white}{\hlc[red1]{t\u{a}i}}\textcolor{white}{\hlc[red1]{pentru}}\textcolor{white}{\hlc[red1]{a}}\textcolor{white}{\hlc[red1]{v\u{a}}}\textcolor{white}{\hlc[red1]{ajuta}}\textcolor{black}{\hlc[red7]{pe}}\textcolor{black}{\hlc[red7]{to\c{t}i}}\textcolor{black}{\hlc[red7]{.}} & \textit{``facebook knows a lot of things about you, most of which are shared with your friends in order to help all of you.''} \\

\#M4 & \textcolor{white}{\hlc[red1]{cei}}\textcolor{white}{\hlc[red1]{mai}}\textcolor{white}{\hlc[red1]{mul\c{t}i}}\textcolor{white}{\hlc[red1]{bani,}}\textcolor{white}{\hlc[red4]{locuitorii}}\textcolor{white}{\hlc[red4]{capitalei}}\textcolor{white}{\hlc[red3]{\^ii}}\textcolor{white}{\hlc[red3]{cheltuie}}\textcolor{white}{\hlc[red3]{\$ne\$}}\textcolor{white}{\hlc[red3]{pe}}\textcolor{white}{\hlc[red3]{m\^incare}}\textcolor{white}{\hlc[red3]{.}} & \textit{``the inhabitants of the capital spend most of their money on food.''}  \\

\#M5 & \textcolor{white}{\hlc[red1]{la}}\textcolor{white}{\hlc[red1]{edi\c{t}ia}}\textcolor{white}{\hlc[red1]{curent\u{a},}}\textcolor{white}{\hlc[red1]{a}}\textcolor{black}{\hlc[red6]{43}}\textcolor{black}{\hlc[red6]{-}}\textcolor{black}{\hlc[red6]{a,}}\textcolor{black}{\hlc[red6]{a}}\textcolor{black}{\hlc[red6]{\$ne\$}}\textcolor{black}{\hlc[red6]{\$ne\$}}\textcolor{black}{\hlc[red6]{de}}\textcolor{black}{\hlc[red6]{\$ne\$}}\textcolor{black}{\hlc[red6]{``\$ne\$}}\textcolor{black}{\hlc[red6]{particip\u{a}}}\textcolor{white}{\hlc[red5]{muzicieni}}\textcolor{white}{\hlc[red5]{din}}\textcolor{white}{\hlc[red2]{15}}\textcolor{white}{\hlc[red2]{\c{t}\u{a}ri,}}\textcolor{white}{\hlc[red2]{iar}}\textcolor{white}{\hlc[red2]{concertele}}\textcolor{white}{\hlc[red3]{se}}\textcolor{white}{\hlc[red3]{desf\u{a}\c{s}oar\u{a}}}\textcolor{white}{\hlc[red3]{\^in}}\textcolor{white}{\hlc[red3]{diferite}}\textcolor{white}{\hlc[red3]{localit\u{a}\c{t}i}}\textcolor{white}{\hlc[red1]{ale}}\textcolor{white}{\hlc[red1]{republicii}}\textcolor{white}{\hlc[red2]{.}} & \textit{``in the current, 43rd edition, of \$ne\$ \$ne\$ \$ne\$ \$ne\$ there are musicians from 15 countries, and the concerts are going to happen in different locations of the republic.''}\\

\#M6 & \textcolor{white}{\hlc[red1]{ma\c{s}ina}}\textcolor{white}{\hlc[red1]{zbur\u{a}toare}}\textcolor{white}{\hlc[red1]{\$ne\$}}\textcolor{white}{\hlc[red5]{a}}\textcolor{white}{\hlc[red5]{companiei}}\textcolor{white}{\hlc[red5]{\$ne\$}}\textcolor{black}{\hlc[red7]{a}}\textcolor{black}{\hlc[red7]{fost}}\textcolor{black}{\hlc[red7]{\^in}}\textcolor{black}{\hlc[red7]{dezvolt\u{a}ri}}\textcolor{black}{\hlc[red10]{\c{s}i}}\textcolor{black}{\hlc[red10]{teste}}\textcolor{black}{\hlc[red10]{timp}}\textcolor{black}{\hlc[red10]{de}}\textcolor{black}{\hlc[red8]{mul\c{t}i}}\textcolor{black}{\hlc[red8]{ani,}}\textcolor{black}{\hlc[red8]{dar}}\textcolor{black}{\hlc[red8]{este}}\textcolor{white}{\hlc[red2]{\^in}}\textcolor{white}{\hlc[red2]{sf\^ir\c{s}it}}\textcolor{white}{\hlc[red2]{aproape}}\textcolor{black}{\hlc[red7]{gata}}\textcolor{black}{\hlc[red7]{.}} & \textit{``the flying car \$ne\$ of the \$ne\$ company has been under development and tests for many years, but it is, finally, almost ready.''} \\

\#M7 & \textcolor{white}{\hlc[red1]{una}}\textcolor{white}{\hlc[red1]{dintre}}\textcolor{white}{\hlc[red1]{cele}}\textcolor{white}{\hlc[red1]{mai}}\textcolor{white}{\hlc[red5]{mari}}\textcolor{white}{\hlc[red5]{b\u{a}nci}}\textcolor{white}{\hlc[red5]{din}}\textcolor{white}{\hlc[red5]{\$ne\$}}\textcolor{white}{\hlc[red3]{\$ne\$}}\textcolor{white}{\hlc[red3]{a}}\textcolor{white}{\hlc[red3]{ajuns}}\textcolor{white}{\hlc[red3]{\^in}}\textcolor{white}{\hlc[red3]{faliment}}\textcolor{white}{\hlc[red1]{.}} & \textit{``one of the greatest banks in \$ne\$ \$ne\$ went bankrupt.''} \\

\#M8 & \textcolor{white}{\hlc[red1]{guvernul}}\textcolor{white}{\hlc[red1]{a}}\textcolor{white}{\hlc[red1]{avizat}}\textcolor{white}{\hlc[red1]{pozitiv,}}\textcolor{white}{\hlc[red3]{\^in}}\textcolor{white}{\hlc[red3]{\c{s}edin\c{t}a}}\textcolor{white}{\hlc[red3]{de}}\textcolor{white}{\hlc[red5]{ast\u{a}zi,}}\textcolor{white}{\hlc[red5]{pachetul}}\textcolor{white}{\hlc[red5]{legislativ}}\textcolor{white}{\hlc[red5]{pentru}}\textcolor{white}{\hlc[red5]{reforma}}\textcolor{white}{\hlc[red3]{fiscal\u{a}}}\textcolor{white}{\hlc[red3]{.}} & \textit{``the government approved, in todays' meeting, 
the legislative package for the tax reform.''} \\

\#M9 & \textcolor{white}{\hlc[red2]{\$ne\$}}\textcolor{white}{\hlc[red2]{din}}\textcolor{white}{\hlc[red2]{\$ne\$}}\textcolor{white}{\hlc[red2]{nu}}\textcolor{white}{\hlc[red2]{va}}\textcolor{white}{\hlc[red1]{comunica}}\textcolor{white}{\hlc[red1]{public}}\textcolor{white}{\hlc[red1]{op\c{t}iunea}}\textcolor{white}{\hlc[red2]{partidului}}\textcolor{white}{\hlc[red2]{privind}}\textcolor{black}{\hlc[red6]{turul}}\textcolor{black}{\hlc[red6]{\$ne\$}}\textcolor{black}{\hlc[red6]{din}}\textcolor{black}{\hlc[red7]{\$ne\$}}\textcolor{black}{\hlc[red7]{dar}}\textcolor{black}{\hlc[red7]{sper\u{a}}}\textcolor{black}{\hlc[red7]{c\u{a}}}\textcolor{black}{\hlc[red6]{la}}\textcolor{black}{\hlc[red6]{3}}\textcolor{black}{\hlc[red6]{iunie}}\textcolor{black}{\hlc[red6]{cet\u{a}\c{t}enii}}\textcolor{white}{\hlc[red5]{se}}\textcolor{white}{\hlc[red5]{vor}}\textcolor{white}{\hlc[red5]{mobiliza}}\textcolor{white}{\hlc[red4]{\c{s}i}}\textcolor{white}{\hlc[red4]{vor}}\textcolor{white}{\hlc[red4]{participa}}\textcolor{white}{\hlc[red4]{activ}}\textcolor{white}{\hlc[red4]{la}}\textcolor{white}{\hlc[red4]{vot,}}\textcolor{white}{\hlc[red4]{a}}\textcolor{white}{\hlc[red4]{declarat}}\textcolor{white}{\hlc[red3]{la}}\textcolor{white}{\hlc[red3]{un}}\textcolor{white}{\hlc[red3]{briefing}}\textcolor{white}{\hlc[red2]{vicepre\c{s}edintele}}\textcolor{white}{\hlc[red1]{forma\c{t}iunii}}\textcolor{white}{\hlc[red1]{\$ne\$}}\textcolor{white}{\hlc[red1]{\$ne\$}}\textcolor{white}{\hlc[red5]{.}} & \textit{``\$ne\$ from \$ne\$ isn't going to publicly announce the party's option regarding \$ne\$ tour in \$ne\$, but they hope that, on June 3rd, the citizens are going to actively participate in the vote, the vice-president of the \$ne\$ \$ne\$ has declared.'' } \\

\#M10 & \textcolor{white}{\hlc[red2]{\$ne\$}}\textcolor{white}{\hlc[red2]{va}}\textcolor{white}{\hlc[red2]{beneficia}}\textcolor{white}{\hlc[red2]{de}}\textcolor{black}{\hlc[red7]{suportul}}\textcolor{black}{\hlc[red7]{exper\c{t}ilor}}\textcolor{black}{\hlc[red10]{}}\textcolor{black}{\hlc[red10]{europeni}}\textcolor{black}{\hlc[red10]{\^in}}\textcolor{black}{\hlc[red10]{procesul}}\textcolor{black}{\hlc[red10]{de}}\textcolor{black}{\hlc[red10]{implementare}}\textcolor{white}{\hlc[red2]{a}}\textcolor{white}{\hlc[red2]{\$ne\$}}\textcolor{white}{\hlc[red2]{\$ne\$}}\textcolor{white}{\hlc[red2]{\c{s}i}}\textcolor{white}{\hlc[red2]{a}}\textcolor{white}{\hlc[red2]{\$ne\$}}\textcolor{white}{\hlc[red1]{\$ne\$}}\textcolor{white}{\hlc[red1]{\$ne\$}}\textcolor{white}{\hlc[red1]{de}}\textcolor{white}{\hlc[red1]{\$ne\$}}\textcolor{white}{\hlc[red2]{\$ne\$}}\textcolor{white}{\hlc[red2]{,}}\textcolor{white}{\hlc[red2]{condi\c{t}ionalit\u{a}\c{t}i}}\textcolor{white}{\hlc[red1]{prev\u{a}zute}}\textcolor{white}{\hlc[red2]{\^in}}\textcolor{white}{\hlc[red2]{capitolul}}\textcolor{white}{\hlc[red2]{privind}}\textcolor{black}{\hlc[red6]{\$ne\$}}\textcolor{black}{\hlc[red6]{de}}\textcolor{black}{\hlc[red6]{\$ne\$}}\textcolor{black}{\hlc[red10]{\$ne\$}}\textcolor{black}{\hlc[red10]{\$ne\$}}\textcolor{black}{\hlc[red10]{\c{s}i}}\textcolor{black}{\hlc[red10]{\$ne\$}}\textcolor{black}{\hlc[red10]{(}}\textcolor{black}{\hlc[red10]{\$ne\$}}\textcolor{black}{\hlc[red10]{)}}\textcolor{black}{\hlc[red10]{a}}\textcolor{black}{\hlc[red10]{\$ne\$}}\textcolor{black}{\hlc[red10]{de}}\textcolor{black}{\hlc[red10]{\$ne\$}}\textcolor{black}{\hlc[red10]{.}} & \textit{``\$ne\$ will benefit from the support of the European experts in the process of implementing the \$ne\$ \$ne\$ and the \$ne\$ \$ne\$ \$ne\$ of \$ne\$ \$ne\$, conditions specified in the chapter regarding \$ne\$ of \$ne\$ \$ne\$ \$ne\$ and \$ne\$ (\$ne\$) of \$ne\$ of \$ne\$.''} \\
\hline 
\end{tabular}
\end{table}

We provide a set of visualizations for Moldavian sentences in Table~\ref{tab_GradCAM_MD}. Sample \#M1 contains a highlighted noun phrase that is a clear indicator of the Moldavian dialect. Indeed, the noun phrase ``cabinetul de mini\c{s}tri'' (the cabinet of ministers) is almost never used in Romanian, where the alternative ``gurvernul'' (the government) is preferred. The noun ``migran\c{t}i'' (migrants) is also unusual in Romanian, the forms ``emigran\c{t}i'' or ``imigran\c{t}i'' being used instead, depending on the context. In samples \#M2, \#M3, \#M4 and \#M6, we can observe a few highlighted words, such as ``m\^{i}na'' (hand), ``f\u{a}c\^{i}ndu'' (doing), ``s\^{i}nt'' (are), ``m\^{i}ncare'' (food) and ``sf\^{i}r\c{s}it'' (end), that reveal the same pattern used only in the Moldavian dialect, namely the use of the vowel ``\^{i}'' inside words. We note that the vowel ``\^{i}'' is used in Romanian only at the beginning of the words. The same sound is spelled by the vowel ``\^{a}'' anywhere else in the word, and the aforementioned Moldavian words would be written as ``m\^{a}na'', ``f\u{a}c\^{a}ndu'', ``m\^{a}ncare'' and ``sf\^{a}r\c{s}it'', respectively. For the verb ``s\^{i}nt'' (are), even the sound is different, the correct Romanian spelling being ``sunt''. In addition, sample \#M3 contains the adverb ``\^{i}mp\u{a}rt\u{a}\c{s}ite'' (distributed), which would likely be replaced by ``partajate'' in Romanian. In sample \#M4, the CNN model focuses on the phrase ``cei mai mul\c{t}i bani'', the distinctive pattern being the placement of this phrase at the beginning of the sentence. In Romanian, the same sentence would be written as follows: ``locuitorii capitalei cheltuie cei mai mul\c{t}i bani pe m\^{a}ncare''. In sample \#M5, we can understand why the network has highlighted the phrase ``ale republicii'' (of the republic) as being a strong indicator for Moldavian, namely because Moldova is considered a republic. Romania was considered a republic only during the communist regime. Hence, example \#M5 does not contain any dialectal patterns. In sample \#M7, the verb phrase ``a ajuns \^{i}n faliment'' (went bankrupt) is distinctive for the Moldavian dialect. In the Romanian dialect, the verb ``a intrat'' would be used instead of ``a ajuns''. Another distinctive verb phase for the Moldavian dialect is present in sample \#M8, namely ``a avizat pozitiv'' (approved). In Romanian, this verb phrase would be replaced by the verb ``a aprobat'', the adverb ``pozitiv'' being implied by the verb. In \#M9, we can observe that ``briefing'' is used to define a short press conference. To express the same concept, a Romanian speaker would use ``declara\c{t}ie de pres\u{a}'' or ``conferin\c{t}\u{a} de pres\u{a}''. Sample \#M9 contains another dialectal pattern. In Moldavian, a political party is typically referred to as ``forma\c{t}iune'', whereas in Romanian, it is referred to as ``partid''. In sample \#M10, the only highlighted dialect pattern that we found interpretable from our perspective is the use of the noun ``condi\c{t}ionalit\u{a}\c{t}i'' (conditions), since we would rather use ``condi\c{t}ii'' in the Romanian dialect. As for the Romanian sentences, we notice that the character-level CNN finds some relevant patterns of the Moldavian dialect. 

We confess that we were not aware of many of the distinctive patterns among the two dialects discovered through the Grad-CAM visualizations. The same applies to our annotators. While both dialects contain about the same words, it seems that differences regarding the preferred synonym to express a certain concept play a very important role in distinguishing among the two dialects. This also explains why people living in Romania or the Republic of Moldova have such a hard time in distinguishing between the dialects. Many of the presented sentences are grammatically and syntactically correct in both dialects, but some word choices in one dialect seem rather unusual in the other dialect. We believe that untrained people can easily mistake such dialectal patterns with the style of the author. We consider that the presented examples elucidate the mystery behind the unreasonable effectiveness of machine learning in Moldavian versus Romanian dialect identification, revealing some interesting dialectal patterns, previously unknown to ourselves. In summary, we consider hypothesis H2 to be true.

\section{Conclusions}
\label{sec_Conclusion}

In this article, we studied dialect identification and related sub-tasks, e.g.~cross-dialect categorization by topic, for an under-studied language, namely Romanian. We experimented with several machine learning models, including novel ensemble combinations, attaining very good performance levels, especially with the ensemble based on model stacking. For example, our ensemble based on stacking attains dialect identification scores above $94\%$ on news articles, above $86\%$ on sentences and up to $70\%$ on tweets. Comparing the ML models with native Romanian or Moldavian speakers, we found a significant performance gap, the average performance of the human annotators being barely above the random chance baseline. For instance, the average accuracy of humans for dialect identification is about $53\%$. In order to find out why ML models attain significantly better results compared to humans, we analyzed Grad-CAM visualizations of the character-level CNN model. The visualizations revealed some interesting dialectal clues, which were too subtle to be observed by the human annotators or by us. We therefore reached the conclusion that the effectiveness of the ML models is explainable in large part through dialectal patterns, although the models can occasionally distinguish the samples based on their subject. In this regard, we believe that the newly-introduced cross-genre setting, in which the models are trained on sentences from MOROCO and tested on tweets collected from a different time span, is more representative for a fair and realistic evaluation.

While our current study is focused on written dialect identification, we aim to address spoken dialect identification in future work. Since the spoken dialect bares more distinctive clues, it will allow us to include other Romanian sub-dialects in our study, e.g.~those spoken in Ardeal or Oltenia regions.

\section*{acknowledgements}
The authors thank reviewers for their valuable feedback leading to significant improvements of the manuscript.





\printendnotes

\bibliography{sample}



\end{document}